\documentclass{article}

\usepackage[preprint]{neurips_2026}

\usepackage[utf8]{inputenc} 
\usepackage[T1]{fontenc}    
\usepackage{hyperref}       
\usepackage{url}            
\usepackage{booktabs}       
\usepackage{amsfonts}       
\usepackage{nicefrac}       
\usepackage{microtype}      
\usepackage{xcolor}         

\usepackage{graphicx}
\usepackage{amsmath}

\usepackage{comment}

\setcitestyle{numbers,square}

\title{A Geometric Perspective on Next-Token Prediction in Large Language Models: Three Emerging Phases}

\author{%
  Gianfranco Lombardo \\
  Department of Engineering and Architecture\\
  University of Parma\\
  Parma, IT 43125 \\
  \texttt{gianfranco.lombardo@unipr.it} \\
  \And
   Giuseppe Trimigno \\
   Department of Engineering and Architecture\\
  University of Parma\\
  Parma, IT 43125 \\
   \texttt{giuseppe.trimigno@unipr.it} \\
   \AND
   Stefano Cagnoni \\
  Department of Engineering and Architecture\\
  University of Parma\\
  Parma, IT 43125 \\
  \texttt{stefano.cagnoni@unipr.it} \\
}

\begin{document}

\maketitle

\begin{abstract}
We investigate the geometry of predictive information across the layers of 
large language models (LLMs). We repurpose representation lenses--learned affine maps trained to predict the next token from intermediate residual streams--as geometric diagnostic tools. Rather than asking \textit{what} the model predicts at each layer, we ask \textit{where} predictive information resides and \textit{how} it evolves across depth. We define at each layer a 
\textit{predictive readout subspace} as the dominant $k$-dimensional singular 
subspace of such a map on the $d$-dimensional residual stream (where $k$ is 
a resolution parameter), and track its trajectory on the Grassmann manifold 
as a similarity profile across layers. The profile is well described by 
unimodal distributions exhibiting a rise, near-plateau, and descent; varying 
$k$ from $1\%$ to $50\%$ of $d$ traces a Pareto frontier between visibility 
and energy retention, yet the same structure emerges at all scales. Across 
eight models from two families (\textsc{Qwen2.5} and \textsc{OLMo2}, 
1B--32B), we identify three geometric phases. Updates are approximately 
orthogonal to the residual stream throughout; what distinguishes the phases 
is their effect on the effective rank, which expands, stabilizes, and 
concentrates. In the first, \textit{Seeding Multiplexing}, feed-forward 
memories and attention layers seed a candidate set in superposition in 
family-specific proportions, with the final token rising as leading candidate 
from $20\%$ to $35\%$ of positions across this phase. In the second, 
\textit{Hoisting Overriding}, updates override existing subspaces to 
concentrate the candidate distribution without expanding the rank. In the 
third, \textit{Focal Convergence}, high-energy low-rank updates write the 
winner into a form aligned with the unembedding direction. Phases~1 and~3 grow slowly with model depth, while 
Phase~2 expands linearly. The additional capacity of deeper 
LLMs is largely absorbed by candidate disambiguation, not by seeding or 
final convergence.
\end{abstract}

\section{Introduction}
\label{sec:intro}
Understanding how large language models process information across layers remains a central problem in mechanistic interpretability. Although the transformer forward pass applies additive updates to a shared residual stream, the roles of individual layers and the structure of the resulting information are still unclear. A key question is geometric: how predictive information for next-token prediction emerges and propagates through depth. A common approach probes this by projecting intermediate residual states into vocabulary space. The LogitLens~\cite{nostalgebraist2020} applies the final unembedding directly, but suffers from \emph{representational drift}: intermediate states are misaligned with the final basis. Superposition~\cite{elhage2021mathematical} further implies that predictive features may be present but hidden in overlapping directions. To address this, subsequent work learns affine corrections to realign intermediate representations before projection. TunedLens~\cite{belrose2023eliciting} trains per-layer maps via distillation; Jump to Conclusions~\cite{yomdin2024jump} learns cross-layer shortcuts; Patchscopes~\cite{ghandeharioun2024patchscopes} unifies these approaches. In all cases, the affine map serves as a \emph{decoding tool} to reveal what the model predicts at each layer (see Section~\ref{sec:background&related}).

\begin{figure}[t]
    \centering
    \includegraphics[width=\linewidth]{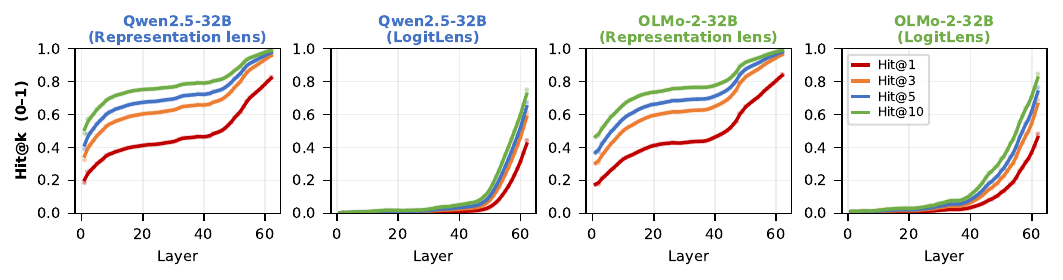}
    \caption{Examples of Hit@$k$ achieved by a representation lens and LogitLens across layers. The representation lens recovers the final token early, while the LogitLens remains nearly blind until the final phase, indicating predictive information is present but hidden in superposition.}
    \label{fig:lens_hit}
\end{figure}

\textbf{Our approach: lenses as geometric probes.} 
Our approach originates from a key observation: Figure~\ref{fig:lens_hit} 
shows layer-wise Hit@$k$ (how often the final token appears in the 
top-$k$ at each layer) for the representation lens and the LogitLens on two 32B models (\textsc{Qwen2.5} and \textsc{OLMo2}). The final token ranks among the top candidates from layer~1 (${\approx}20\%$ Hit@1 over a ${\sim}100$k vocabulary), while the LogitLens remains nearly blind across both architectures. Consequently, predictive information is present from the earliest layers but hidden in superposition, misaligned with the unembedding direction. LoRA-Lens~\cite{trimigno2026loralens} shows that only $3$--$6\%$ of the hidden dimensionality suffices to match full-rank fidelity: the lens reads the predictive signal by projecting onto the low-dimensional subspace where it is concentrated. This leads us to repurpose affine maps as \emph{geometric probes}: 
rather than asking \textit{what} the model predicts at each layer, we ask 
\textit{where} predictive information resides in the residual stream and 
\textit{how} that location evolves across depth. 

This paper makes two main contributions: 

\noindent\textbf{(a) A framework for tracking predictive geometry across depth.} 
We train affine maps following ~\citep{belrose2023eliciting} on SlimPajama~\citep{slimpajama} (226M tokens) for eight models from two families (\textsc{Qwen2.5}~\cite{yang2025qwen3} and \textsc{OLMo2}~\cite{olmo20242}, 1B--32B). We decompose each map via SVD and define the \textit{predictive readout subspace} as the top-$k$ right singular directions acting on the residual stream. We track its trajectory on $\mathrm{Gr}(k,d)$ via a layer-wise similarity profile, revealing structural changes across depth. 

\noindent\textbf{(b) Three emerging geometric phases of next-token prediction.} 
Across models and scales, the similarity profile exhibits a rise, plateau, and descent, defining three phases. We characterize them jointly: \textit{(1) Seeding Multiplexing}, where candidates are seeded in superposition and rank saturates; \textit{(2) Hoisting Overriding}, where updates concentrate the distribution without expanding rank; and \textit{(3) Focal Convergence}, where low-rank high-energy updates write the final token. Phases~1 and~3 scale weakly with depth, while Phase~2 grows linearly, indicating that additional capacity is primarily used for candidate disambiguation.

\section{Background and Related Work}
\label{sec:background&related}
\subsection{Background}
\label{sec:background}

\textbf{The residual stream and transformer computation.}
A large language model with $L$ layers processes an input
token sequence $(x_1, \ldots, x_T)$ by maintaining, at each position $t$, a
residual stream $h_\ell \in \mathbb{R}^d$ that is updated additively across
layers. 
%
%
Each layer $\ell \in \{1, \ldots, L\}$ applies two
sub-modules in sequence, a multi-head self-attention (MHA) sublayer and a
feed-forward network (FFN) sublayer, both contributing additive
updates to the residual stream:
\begin{equation}
    h_\ell = h_{\ell-1} + \Delta_\ell^{\mathrm{MHA}} + \Delta_\ell^{\mathrm{FFN}},
    \label{eq:residual_update}
\end{equation}
where $\Delta_\ell^{\mathrm{MHA}}, \Delta_\ell^{\mathrm{FFN}} \in \mathbb{R}^d$
denote the attention and feed-forward contributions at layer $\ell$, each of
which may incorporate layer normalization depending on the architecture. The
additive structure in Eq.~\ref{eq:residual_update} ensures that information
is preserved and incrementally refined across depth: each layer writes a
residual increment to the shared stream, which accumulates monotonically. The
next-token distribution is obtained at the final layer by applying a layer
normalization followed by the unembedding matrix $W_U \in \mathbb{R}^{|V|
\times d}$:
\begin{equation}
    p(x_{t+1} \mid x_{\leq t})
    = \mathrm{softmax}\!\left(W_U\,\mathrm{LN}(h_L)\right),
\end{equation}
where $\mathrm{LN}(\cdot)$ denotes layer normalization. The composition $W_U
\circ \mathrm{LN}$ defines the model's final linear readout from the terminal
residual state into logit space.

\textbf{Representation lenses.}

\citet{alain2017understanding} probe intermediate layers with linear readouts to assess decodable information. Building on this, the LogitLens~\cite{nostalgebraist2020} applies the final readout $W_U \circ \mathrm{LN}$ directly to intermediate states $h_\ell$, but is limited by \emph{representational drift}, as intermediate representations may be misaligned with $W_U$. Representation lenses address this by learning an affine correction:
\begin{equation}
    \tilde{h}_\ell = (I + A_\ell)\,h_\ell + b_\ell, 
    \qquad 
    z_\ell = W_U\,\mathrm{LN}(\tilde{h}_\ell),
\end{equation}
where $(A_\ell, b_\ell)$ are trained to align intermediate predictions with the final output via KL distillation. The identity term ensures the LogitLens is recovered when $A_\ell=0$, making $A_\ell$ the key realignment component. This framework underlies methods such as TunedLens~\cite{belrose2023eliciting}, Jump to Conclusions~\cite{yomdin2024jump}, and Patchscopes~\cite{ghandeharioun2024patchscopes}, which differ in scope but share the same affine correction principle.

\subsection{Related Work}

\textbf{Mechanistic interpretability and the residual stream.}
A central line of work in transformer interpretability analyzes components by decomposing the forward pass into circuits over a shared residual stream. \citet{elhage2021mathematical} formalize this framework and introduce superposition, where features are encoded in overlapping directions. Isolating predictive information thus requires identifying its low-dimensional structure. \citet{elhage2023privileged} study preferred bases of the residual stream, while causal approaches~\citep{ameisen2025circuit} intervene on components. In contrast, we analyze how the geometry of predictive information evolves across depth.

\textbf{Geometric and linear structure of internal representations.}
Recent work explores transformer representations exhibit interpretable geometric structure. \citet{park2024linearrepresentationhypothesisgeometry} formalize this via the Linear Representation hypothesis, suggesting high-level features are encoded as linear directions in activation space. Empirical results show this structure emerges from the training objective rather than architectural constraints~\citep{nanda-etal-2023-emergent}. \citet{shai2024transformers} provide evidence the residual stream encodes geometric objects corresponding to belief states, reflecting the probabilistic structure of the task. Additionally, \citet{gurnee2026modelsmanipulatemanifoldsgeometry} show transformers construct and manipulate low-dimensional manifolds within the residual stream, motivating our Grassmann manifold trajectory analysis.

\textbf{Depth-dependent computation and phase structure.}
Empirical evidence shows that transformer layers are not functionally homogeneous: \citet{geva-etal-2021-transformer} demonstrate that feed-forward sublayers act as key-value memories with depth-varying associations, while \citet{meng2022locating} show that factual knowledge is mainly localized in middle layers.
\citet{aghajanyan-etal-2021-intrinsic} show that fine-tuning has low intrinsic dimensionality, implying relevant information lies in a low-dimensional subspace, consistent with our finding that predictive directions occupy a small part of the residual stream. Finally, \citet{bruno2026a} identify three dynamical phases in a mean-field transformer: an alignment phase with spectral collapse, an intermediate heat phase, and a slow pairing phase. This parallels our three-phase decomposition, though their analysis considers encoder-only models in the large-token limit, while ours focuses on autoregressive decoder-only models and the geometry of predictive correction maps. We compare our results with \citet{bruno2026a} findings in the final discussion.

\section{Methodology}
\label{sec:method}

\subsection{Predictive Readout Subspace and Grassmann trajectory}
\label{sec:subspace}

A representation lens at layer $\ell$ is defined by a correction matrix $A_\ell \in \mathbb{R}^{d \times d}$ and a bias $b_\ell \in \mathbb{R}^d$.
Since the bias only acts as a global translation, we focus on $A_\ell$, which captures directional corrections to the residual stream. We decompose it via SVD and define the predictive readout subspace from its top singular directions:
\begin{equation}
\begin{aligned}
A_\ell &= U_\ell \Sigma_\ell V_\ell^\top, \quad &
\mathcal{S}_\ell^{(k)} &= \mathrm{span}\!\left(V_\ell^{(k)}\right).
\end{aligned}
\label{eq:svd_subspace}
\end{equation}

Here, the singular values in $\Sigma_\ell$ quantify transformation strength along each direction, and the top right singular vectors identify where the lens applies the strongest corrections. Let $A_\ell^{(k)}$ denote the rank-$k$ truncation of $A_\ell$, with 
corresponding corrected state $\tilde{h}_\ell^{(k)} = (I + 
A_\ell^{(k)})\,h_\ell + b_\ell$ and prediction $z_\ell^{(k)} = 
W_U\,\mathrm{LN}(\tilde{h}_\ell^{(k)})$.
If:
{\small\begin{equation}
    \mathbb{E}_{h_\ell}\!\left[
        \mathrm{KL}\!\left(
            \mathrm{softmax}(z_\ell)
            \,\Big\|\,
            \mathrm{softmax}(z_\ell^{(k)})
        \right)
    \right] \leq \varepsilon
    \label{eq:kl_bound}
\end{equation}}
holds for small $\varepsilon$, the predictive mapping is well-approximated by the projection of $h_\ell$ onto 
$\mathcal{S}_\ell^{(k)}$. This does not imply that $h_\ell$ is itself 
low-rank, but only that predictive information is compressible under linear readout. Empirically, LoRA-Lens~\citep{trimigno2026loralens} shows that ranks between 
$3\%$ and $10\%$ of hidden dimensionality $d$ suffice to match full performance, indicating that a few singular directions dominate. In light of this low-rank approximation, we define the predictive readout subspace $\mathcal{S}_\ell^{(k)}$ as a $k$-dimensional subspace of $\mathbb{R}^d$, viewed as a point on the Grassmann manifold $\mathrm{Gr}(k, d)$. In such a manifold, relations between subspaces are characterized by principal angles, computed from the singular values of $Q_{\mathcal{A}}^\top Q_{\mathcal{B}}$, which capture their relative geometry and enable similarity measures. The sequence $\{\mathcal{S}_\ell^{(k)}\}_{\ell=1}^L$ defines a trajectory on $\mathrm{Gr}(k, d)$, tracking how the dominant 
directions of predictive readout shift across the layers.

\begin{figure}[t]
    \centering
    \includegraphics[width=\linewidth]{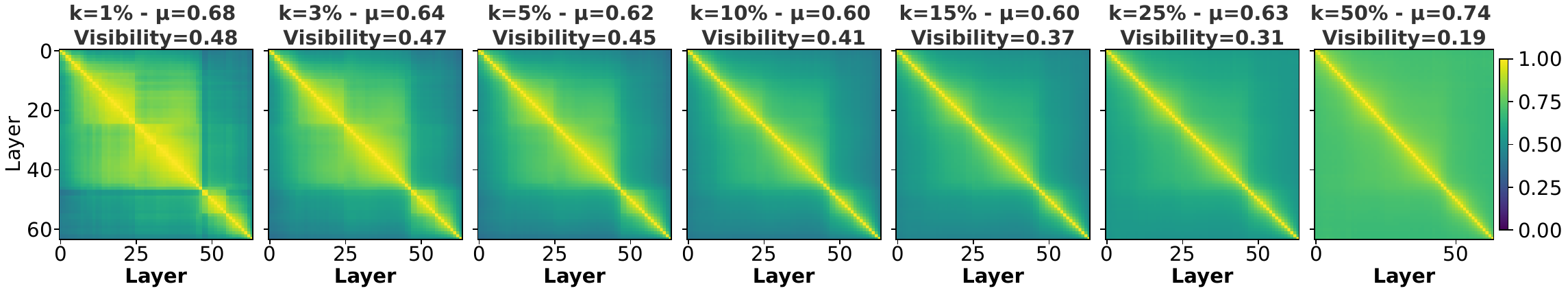}
    \caption{Pairwise RSS similarity matrices $\mathcal{M}^{(k)}$ for 
    \textsc{OLMo2-32B} at truncation levels $k$. 
    }
    \label{fig:sim_matrices}
\end{figure}

\begin{figure}[t]
    \centering
    \includegraphics[width=0.9\linewidth]{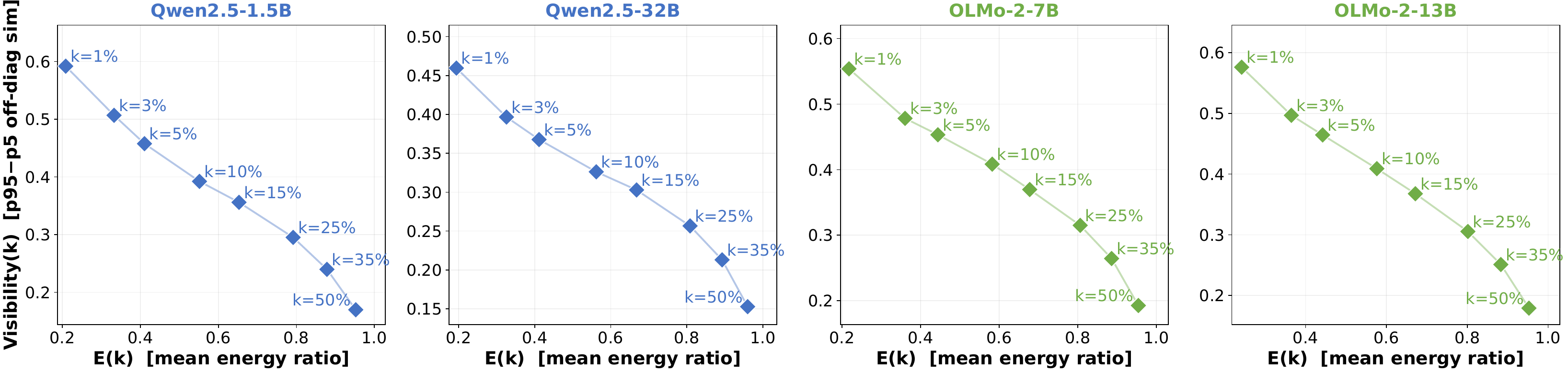}
    \caption{Examples of Pareto frontier between Visibility@k and $E^{(k)}$. No single 
    $k$ dominates.}
    \label{fig:pareto_frontier}
\end{figure}

\textbf{Readout Subspace Similarity (RSS). }To characterize how the trajectory $\{\mathcal{S}_\ell^{(k)}\}$ evolves
across layers, we measure the alignment between consecutive subspaces on
$\mathrm{Gr}(k, d)$ via their principal angles. We adopt a normalized
similarity score, the \textit{Readout Subspace Similarity (RSS)}, defined
as the mean cosine of principal angles between $\mathcal{S}_\ell^{(k)}$
and $\mathcal{S}_{\ell+1}^{(k)}$. This yields a score in $[0, 1]$ with
immediate interpretation ($\mathrm{RSS}_\ell^{(k)} = 1$: identical
subspaces; $\mathrm{RSS}_\ell^{(k)} = 0$: orthogonal subspaces), and is
invariant to $\|A_\ell\|_F$, isolating geometric reorientation from
scaling effects. Let $Q_\ell^{(k)}, Q_{\ell+1}^{(k)} \in \mathbb{R}^{d
\times k}$ be orthonormal bases for $\mathcal{S}_\ell^{(k)}$ and
$\mathcal{S}_{\ell+1}^{(k)}$, obtained via QR decomposition of the
respective top-$k$ right singular vectors. We define:
{\small\begin{equation}
    \mathrm{RSS}_\ell^{(k)}
    = \frac{1}{k}\sum_{r=1}^{k} \cos\theta_r^{(\ell)}
    \label{eq:vsim}
\end{equation}}
where $\{\theta_r^{(\ell)}\}_{r=1}^k$ are the principal angles between
$\mathcal{S}_\ell^{(k)}$ and $\mathcal{S}_{\ell+1}^{(k)}$. Tracking
$\mathrm{RSS}_\ell^{(k)}$ across layers yields the \textit{similarity
profile} $\{\mathrm{RSS}_\ell^{(k)}\}_{\ell=1}^{L-1}$, a layer-wise
characterization of how the predictive readout geometry evolves through
the forward pass.

\begin{figure}[t]
    \centering
    \includegraphics[width=0.9\linewidth]{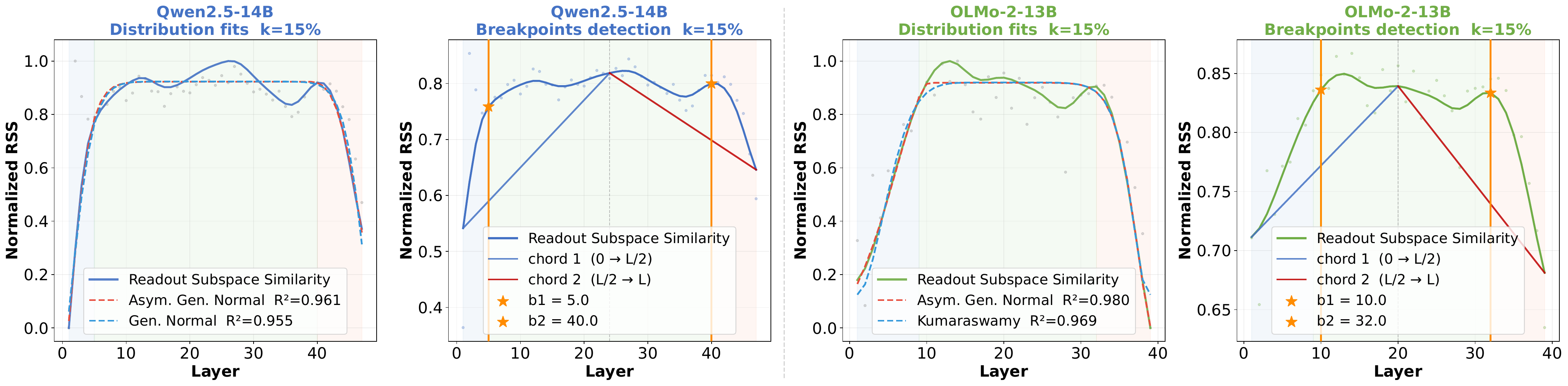}
    \caption{Examples of unimodal fits (left) and breakpoint detection (right) on the RSS profile @k=15\%. Despite small local deviations, chord-based detection remains robust.}
    \label{fig:example_breakpoints}
\end{figure}

\begin{figure}[t]
    \centering
    \includegraphics[width=0.9\linewidth]{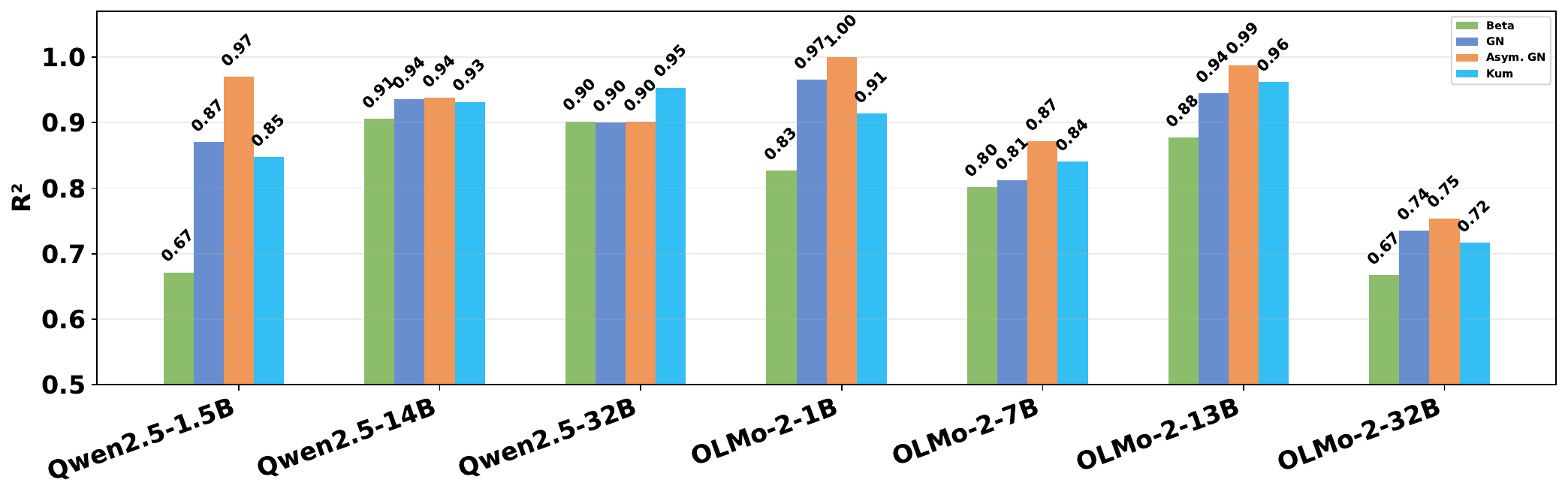}
    \caption{Distributional fits to RSS profiles $\{s_\ell^{(k)}\}$ at 
    $k=50\%$. High $R^2$ values for unimodal distributions supports the 
    unimodality assumption and the tripartite decomposition.}
    \label{fig:unimodal_fit}
\end{figure}

\subsection{Phase boundaries detection}
\label{sec:resolution}
 
\textbf{Experimental setup.}
We train representation lenses following the TunedLens 
formulation~\citep{belrose2023eliciting} on 
SlimPajama~\citep{slimpajama} ($226$M tokens) for eight models from two 
architecturally distinct families (\textsc{Qwen2.5} and \textsc{OLMo2}, 
1B--32B). See Appendix~\ref{app:lens_training} for the full dataset and training details. For each model, we compute $\{\mathcal{S}_\ell^{(k)}\}_{\ell=1}^L$ 
at truncation levels $k \in \{1, 3, 5, 10, 15, 25, 35, 50\}\%$ of $d$, 
deriving: (i) the similarity profile $\{s_\ell^{(k)}\}$ and (ii) the full 
pairwise similarity matrix $\mathcal{M}^{(k)}$, where 
$\mathcal{M}^{(k)}_{ij}$ denotes the RSS between subspaces at layers $i$ 
and $j$.
 
\textbf{Resolution-Scale Analysis and Pareto Frontier.}
To quantify geometric discriminability at resolution $k$, we define:
\[
\mathrm{Visibility@k}
\;=\;
p_{95}\!(\mathcal{M}^{(k)})
-
p_{5}\!(\mathcal{M}^{(k)}),
\]
where $p_{95}$ and $p_5$ are the 95th and 5th percentiles of the 
off-diagonal entries of $\mathcal{M}^{(k)}$. This percentile-based spread 
is robust to outliers: high values indicate well-separated subspaces; low 
values indicate convergence toward a degenerate configuration. Figure~\ref{fig:sim_matrices} shows the pairwise similarity matrices 
$\mathcal{M}^{(k)}$ for \textsc{OLMo2-32B} across all truncation levels, 
illustrating how Visibility@K decreases as $k$ increases. Corresponding results for all models are provided in Appendix~\ref{app:sim_matrices}. We contrast 
this with the mean retained spectral energy:
{\small
\[
E^{(k)} = \frac{1}{L}\sum_{\ell=1}^{L}
\frac{\sum_{i=1}^{k}\sigma_{\ell,i}^{2}}{\sum_{i=1}^{d}\sigma_{\ell,i}^{2}}
\]}
which quantifies how much of the lens transformation is captured by the 
top-$k$ singular directions. Increasing $k$ monotonically improves $E^{(k)}$ but reduces 
$\mathrm{Visibility@k}$: as $k \to d$, all subspaces coincide with 
$\mathbb{R}^d$ and pairwise similarity collapses to $1$. This defines a Pareto trade-off between fidelity and discriminability, making $k$ a \emph{resolution parameter}.  As shown in Figure~\ref{fig:pareto_frontier} (see Appendix~\ref{app:pareto} for all models), intermediate values $k \in \{5\%, 10\%, 15\%\}$ balance spectral energy and geometric contrast, while smaller $k$ are noisier and larger $k$ oversmooth. Our phase detection therefore aggregates results across this regime. Robustness to other $k$ values is discussed in Section~\ref{sec:results}.

\textbf{Unimodality Assumption.}
RSS profiles ${s_\ell^{(k)}}$ exhibit a broad, plateau-like unimodal shape across models and resolutions (Figure~\ref{fig:example_breakpoints}; Appendix~\ref{app:unimodality_phase_detection}). This suggests they can be described by a rise, near-plateau, and descent, motivating a three-region decomposition and a two-breakpoint model.
To validate this assumption, we fit four parametric families across models and $k$: Beta, generalized normal (GNorm), asymmetric GNorm (AGNorm), and Kumaraswamy. GNorm and AGNorm enforce unimodality, while Beta and Kumaraswamy allow more flexible shapes. If the latter do not consistently improve fit, this supports the adequacy of the unimodal assumption. Results are reported in Section~\ref{sec:results}.

\textbf{Breakpoints detection.}
Given the unimodal similarity profile $\{s_\ell^{(k)}\}_{\ell=1}^{L-1}$, we identify two breakpoints $b_1 < b_2$ that partition depth into three phases via a two-stage Kneedle criterion~\citep{satopaa2011finding}, applied separately to each half of the profile. This chord-based method is assumption-light, avoids derivatives, and is robust to local deviations from unimodality (Figure~\ref{fig:example_breakpoints}), as it relies on global geometry. All computations use a Gaussian-smoothed profile with $\sigma = 2$
(formal procedure in Appendix \ref{app:unimodality_phase_detection}). 
The midpoint split follows from unimodality, placing one transition in each half. Breakpoint estimates may vary by $1$--$3$ layers across values of $k$ due 
to residual noise. We aggregate over the intermediate regime 
$k \in \{5\%, 10\%, 15\%\}$ using the mode, or the median when all three estimates are distinct. 

\subsection{Functional Interpretation Setup}
\label{sec:functional_setup}
To characterize phase behavior, we collect residual stream activations using a held-out validation subset of SlimPajama~\citep{slimpajama} (16M tokens across six domains: CommonCrawl, C4, ArXiv, GitHub, StackExchange, Wikipedia). Using validation data prevents evaluation on optimized data, ensuring metrics reflect intrinsic model computation rather than artifacts of lens training, and avoids spurious correlations. The multi-domain mix further reduces domain bias. Activations are collected in batches: for each layer $\ell$ and token $t$ we record $h_\ell^{(t)} \in \mathbb{R}^d$ and the total update $\Delta_\ell^{(t)} = \Delta_\ell^{\mathrm{MHA},(t)} + \Delta_\ell^{\mathrm{FFN},(t)}$. Metrics are averaged over tokens and batches. We compute four measures: (i) effective rank of hidden states, (ii) mean $|\cos(\Delta_\ell, h_{\ell-1})|$, (iii) lens Hit@$k$, and (iv) the norm ratio $|\Delta_\ell^{\mathrm{FFN}}|/|\Delta_\ell^{\mathrm{MHA}}|$.

\begin{figure}[t]
    \centering
    \includegraphics[width=0.95\linewidth]{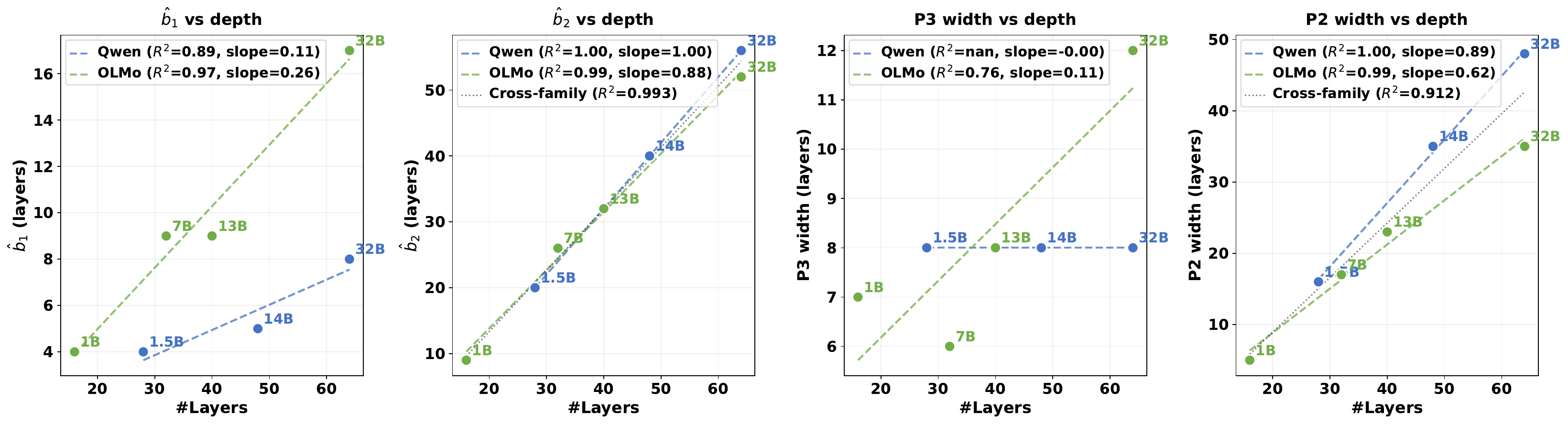}
    \caption{Scaling structure of the three phases across model depths. From     left to right: $\hat{b}_1$ vs depth, $\hat{b}_2$ vs depth, Phase~3 width vs depth, and Phase~2 width vs depth.}
    \label{fig:scaling}
\end{figure}

\begin{figure}[t]
    \centering
    \includegraphics[width=0.9\linewidth]{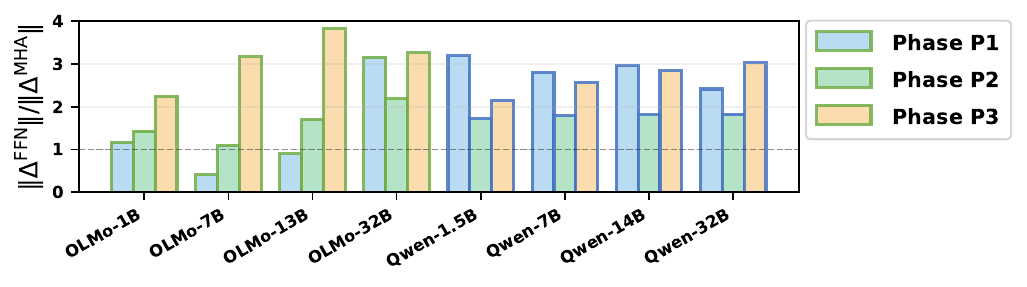}
    \caption{$\|\Delta_\ell^{\mathrm{FFN}}\|/\|\Delta_\ell^{\mathrm{MHA}}\|$
    ratio across phases per model.}
    \label{fig:mlp_mha}
\end{figure}

\begin{figure}[t]
    \centering
    \includegraphics[width=0.9\linewidth]{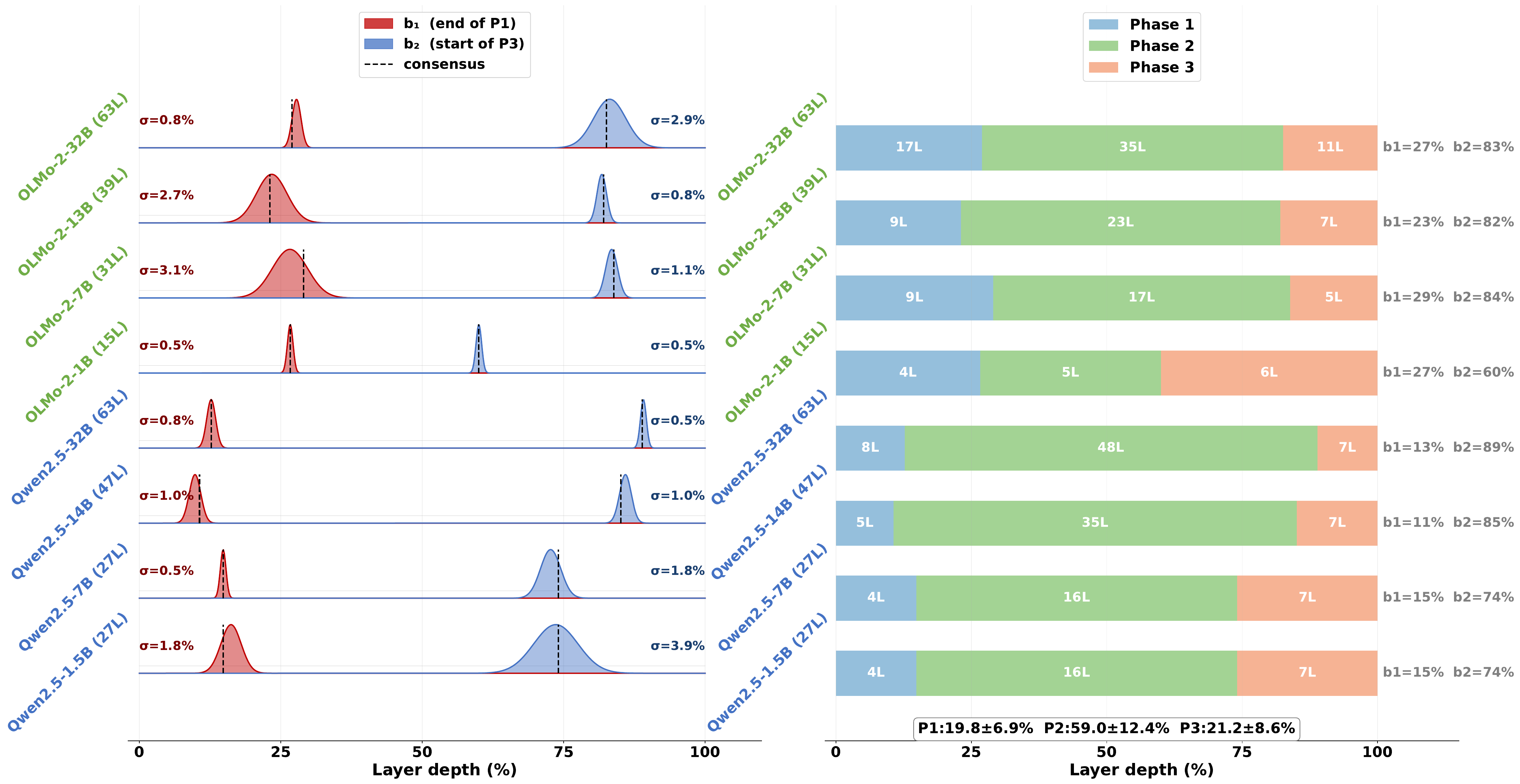}
    \caption{Left: breakpoint robustness for all eight 
    models across all subspace resolutions $k 
    \in \{1\%,.., 50\%\}$. Right: consensus phase decomposition with depth-normalized layers. }
    \label{fig:comparison}
\end{figure}

\begin{figure}[t]
    \centering
    \includegraphics[width=\linewidth]{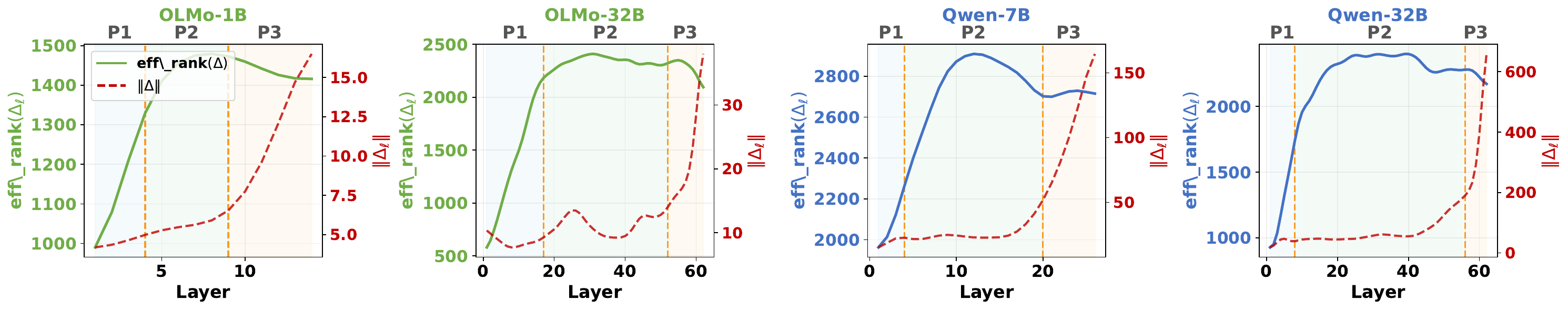}
    \caption{Rank of the total update $\Delta_\ell$ (solid) and its norm
    (dashed). All models in
    Appendix~\ref{app:geometric_characterization}.}
    \label{fig:delta_collapse}
\end{figure}

\section{Results}
\label{sec:results}
 
\subsection{Phase detection}
\label{sec:results_unimodal}
 
\textbf{Validation of the Unimodality Assumption.}
We report the distributional fits from Section~\ref{sec:method} as validation of the unimodality assumption. Across models and $k$, fits achieve high $R^2$ ($\geq0.9$ for most models, minimum 0.75; Figure~\ref{fig:unimodal_fit}), showing profiles are well captured by simple unimodal forms. This holds across all $k$: small $k$ yields noisier profiles, while large $k$ flattens them, reducing geometric contrast (Figure~\ref{fig:sim_matrices}), yet unimodal fits remain strong even at $k=50\%$. Importantly, more flexible families (Beta, Kumaraswamy) do not outperform unimodal ones (GNorm, AGNorm). When performance is similar, extra flexibility is unused; when unimodal fits are better, no additional structure is present. In both cases, RSS profiles are well described as unimodal, exhibiting a rise, plateau, and descent, supporting the two-breakpoint, three-phase model.

\textbf{Phase Identification and Robustness.}
Applying the consensus procedure across eight models, we identify two breakpoints $\hat{b}_1,\hat{b}_2$ partitioning the RSS into three phases (Figure~\ref{fig:comparison}). Boundaries are consistent across architectures. On average, Phase~1 spans $19.8\% \pm 6.9\%$, Phase~2 $59.0\% \pm 12.4\%$, and Phase~3 $21.2\% \pm 8.6\%$, with the intermediate phase dominating.
Comparing with breakpoints at out-of-regime $k \in {1\%,3\%,25\%,35\%,50\%}$ shows tightly concentrated estimates ($\sigma \leq 3.9\%$; Figure~\ref{fig:comparison}), indicating stable boundaries. This supports that the tripartite structure reflects intrinsic geometric organization rather than artifacts of $k$ or the detection method.

\textbf{Scaling Laws of Phase Boundaries.}
Figure~\ref{fig:scaling} shows how $\hat{b}_1$, $\hat{b}_2$, and phase widths scale with depth the number of layers $L$. The onset of Phase~3 ($\hat{b}_2$) scales linearly with $L$ (Qwen: $R^2=1.00$, slope $=1.00$; OLMo: $R^2=0.99$, slope $=0.88$; cross-family $R^2=0.993$). Phase~2 width follows a similar trend (Qwen: $R^2=1.00$, slope $=0.89$; OLMo: $R^2=0.99$, slope $=0.62$; cross-family $R^2=0.912$), indicating proportional expansion with depth. In contrast, $\hat{b}_1$ shows a slower scaling (Qwen: $R^2=0.89$, slope $=0.11$; OLMo: $R^2=0.97$, slope $=0.26$), suggesting Phase~1 occupies a nearly fixed number of layers. Phase~3 width is also almost constant (Qwen: slope $=0.00$; OLMo: $=0.11$).
Overall, as depth increases, Phase~2 expands while Phases~1 and 3 remain roughly constant.
 
\begin{figure}[t]
    \centering
    \includegraphics[width=\linewidth]{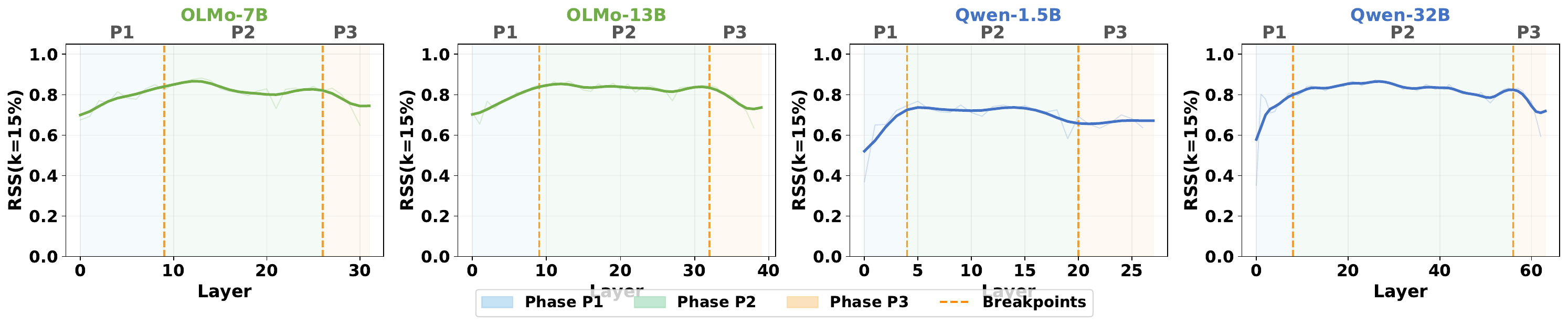}
    \caption{RSS profiles $s_\ell^{(k=15\%)}$ for all eight models across 
    the two architectural families, with consensus breakpoints $\hat{b}_1$ 
    and $\hat{b}_2$ indicated. All models in Appendix \ref{app:geometric_characterization}.}
    \label{fig:phase_overview}
\end{figure}
 
\begin{figure}[t]
    \centering
    \includegraphics[width=\linewidth]{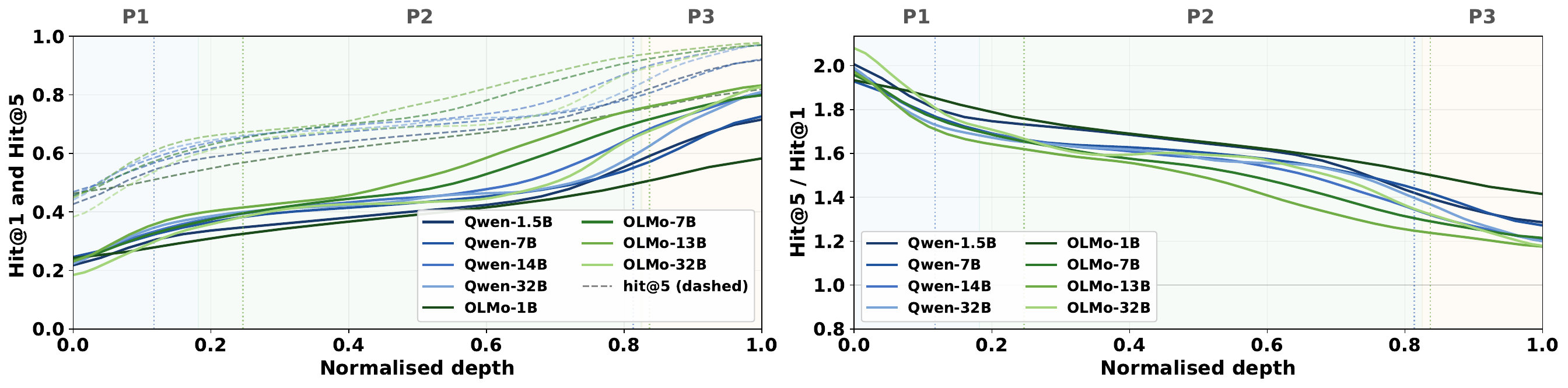}
    \caption{Left: Hit@1 (solid) and Hit@5 
    (dashed) across normalised depth. Right: Hit@5/Hit@1 ratio. 
    Both metrics evolve continuously across all three phases, reflecting a disambiguation process.}
    \label{fig:candidates}
\end{figure}

\subsection{Geometric Characterization of the Three Phases}
\label{sec:results_characterization}

Figure~\ref{fig:phase_overview} shows RSS profiles with consensus
breakpoints. Phase~1 exhibits increasing alignment as predictive geometry
stabilises, Phase~2 maintains a near-plateau, and Phase~3 shows a decline
as the predictive subspace reorients toward the unembedding direction.
Figure~\ref{fig:candidates} overlays representation lens accuracy: Hit@1
increases monotonically while Hit@5/Hit@1 decreases smoothly across all
phases (8/8 models). Already at layer~1, Hit@1 is $\approx 0.28$--$0.33$
and Hit@5/Hit@1 $\approx 2.0$, indicating the final token is among the top
candidates from the earliest layers but not yet dominant.

\begin{figure}[t]
    \centering
    \includegraphics[width=0.9\linewidth]{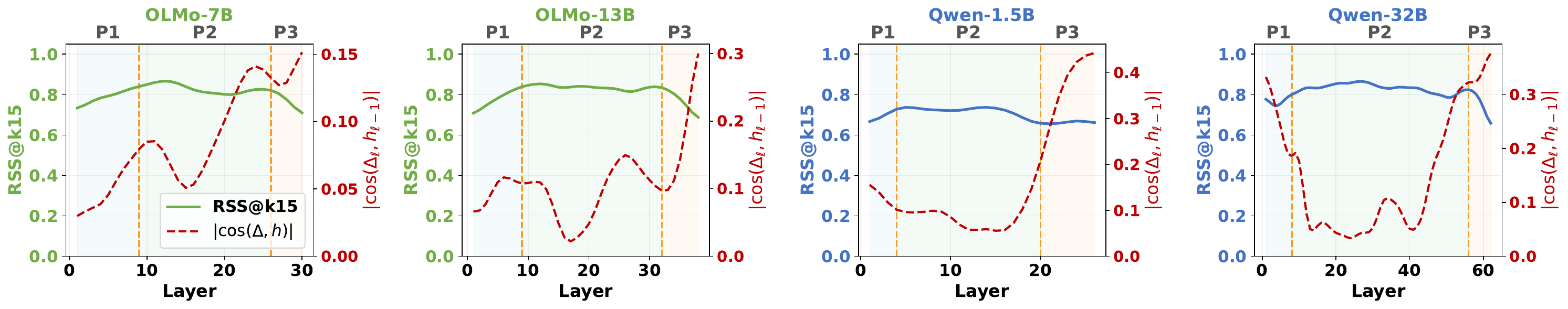}
    \caption{RSS profile $s_\ell^{(k=15\%)}$ (solid, left axis) and
    $|\cos(\Delta_\ell, h_{\ell-1})|$ (dashed red, right axis). High RSS
    $\leftrightarrow$ low cosine (Phase~2); low RSS $\leftrightarrow$ high
    cosine (Phase~3). See Appendix~\ref{app:geometric_characterization} for
    all models.}
    \label{fig:bridge}
\end{figure}

\textbf{A general property: near-orthogonality of updates.}
Layer updates $\Delta_\ell$ are nearly orthogonal to the residual stream
$h_{\ell-1}$ throughout (Figure~\ref{fig:bridge}), consistent with
superposition~\citep{elhage2021mathematical}, so phases differ in their
geometric effect on the predictive subspace rather than in their alignment
with it. $|\cos(\Delta_\ell, h_{\ell-1})|$ is lowest in Phase~2 (7/8
models) and highest in Phase~3 (8/8 models), yielding a consistent
anti-correlation with the RSS profile visible in Figure~\ref{fig:bridge}.

\begin{figure}[t]
    \centering
    \includegraphics[width=\linewidth]{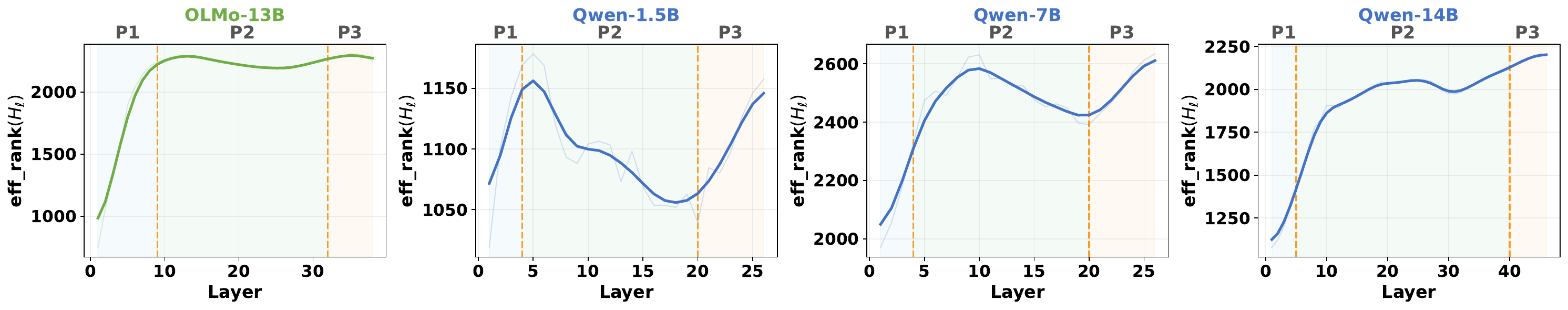}
    \caption{Average effective rank of hidden states across 16M tokens.
    All models in Appendix~\ref{app:geometric_characterization}.}
    \label{fig:rank_saturation}
\end{figure}

\textbf{Effective rank of hidden states.}
Figure~\ref{fig:rank_saturation} shows effective rank increasing through
Phase~1, peaking near $\hat{b}_1$, and stabilising in Phase~2.
\textsc{Qwen2.5-1.5B} is an exception, showing a modest Phase~2 dip
($\approx 80$ dimensions), which we attribute to its constrained capacity
($d = 1536$, 27 layers). Figure~\ref{fig:delta_collapse} shows update rank
peaking in Phase~2 (8/8 models) and declining in Phase~3 (7/8), while
$\|\Delta_\ell\|$ grows monotonically with sharp acceleration in Phase~3
(up to $10\times$), indicating concentration on fewer high-energy
directions.

\textbf{FFN/MHA update norm ratio.}
Figure~\ref{fig:mlp_mha} shows
$\|\Delta_\ell^{\mathrm{FFN}}\|/\|\Delta_\ell^{\mathrm{MHA}}\|$ across
phases. In Phase~1, families differ: \textsc{Qwen2.5} is FFN-dominated
(ratio $2.4$--$3.2$) while \textsc{OLMo2} shows more balanced updates due to its parallel design~\citep{olmo20242}, yet both
converge to the same geometry at $\hat{b}_1$. Contributions are more
balanced in Phase~2, and FFN dominates across both families in Phase~3,
suggesting activation of distributed memories to write the final prediction
into the residual stream.

\subsection{Functional Interpretation of the Three Phases}
\label{sec:interpretation}
The metrics jointly reveal a three-phase functional organization of the
forward pass with distinct geometric and predictive signatures, which we
now interpret and name.

\textbf{Phase~1 -- \textit{Seeding Multiplexing}.}
Nearly orthogonal updates add new directions, driving $H_\ell$ to peak rank
while the predictive subspace reorients (rising RSS); the lens recovers the
final token among several candidates (Hit@5/Hit@1 $\approx 1.74$--$1.87$),
though LogitLens remains ineffective (Hit@1 $\approx 0.01$), indicating
superposed representations. The breakpoint $\hat{b}_1$ scales weakly with
depth (slope $\leq 0.26$), suggesting a fixed cost. Despite architectural
differences (FFN-dominant in \textsc{Qwen2.5}, attention-comparable in
\textsc{OLMo2}), both reach the same geometry. We interpret this phase as
\textit{Seeding Multiplexing}: candidates are seeded in the residual stream
and encoded in superposition until representational capacity saturates.

\textbf{Phase~2 -- \textit{Hoisting Overriding}.}
Hidden-state rank is stable and $|\cos(\Delta_\ell, h_{\ell-1})|$ is
minimal (7/8 models), indicating maximally orthogonal updates when the
predictive subspace is most stable. Hit@1 rises to $\approx 0.56$ and
Hit@5/Hit@1 decreases, showing progressive concentration, while LogitLens
remains low ($\approx 0.03$--$0.10$), implying disambiguation within the
subspace without explicit decoding. FFN and MHA contributions are balanced.
We interpret this phase as \textit{Hoisting Overriding}: one candidate is
progressively lifted above the others (\textit{hoisting}) by overriding
competing subspaces (\textit{overriding}), without altering the established
predictive geometry.

\textbf{Phase~3 -- \textit{Focal Convergence}.}
$|\cos(\Delta_\ell, h_{\ell-1})|$ is maximal (8/8), the rank of
$\Delta_\ell$ decreases (7/8), and update norms grow sharply (up to
$10\times$), concentrating on few high-energy directions within the
predictive subspace. The Hit@5/Hit@1 ratio drops ($\approx 1.2$--$1.5$)
while LogitLens Hit@1 rises ($\approx 0.30$--$0.50$), approaching the
representation lens, indicating the final token is directly readable. FFN
dominates across all models. We interpret this phase as \textit{Focal
Convergence}: updates concentrate on a single target and the candidate
distribution collapses to a dominant prediction.

\section{Discussion}
\label{sec:discussion}
\textbf{Comparison with Mean-Field Theory.}
\label{sec:meanfield}
The tripartite structure of \citet{bruno2026a} for encoder-only models aligns with ours for decoder-only ones: \emph{alignment} matches Phase~1 (rank saturation and subspace stabilisation), \emph{heat} parallels Phase~2 (stable subspace with orthogonal, concentrating updates), and \emph{pairing} contrasts with Phase~3, which is abrupt and directional. This reflect dependence on the model’s objective since encoder-only models does not generate tokens. This agreement suggests a robust structural property in transformer architectures.

\textbf{Conclusions.}
\label{sec:conclusions}
We propose a geometric framework for analyzing next-token prediction, using representation lenses as diagnostic tools. By decomposing the correction matrix via SVD and tracking the predictive subspace on the Grassmann manifold, we identify a consistent three-phase structure across models: \textit{Seeding Multiplexing}, \textit{Hoisting Overriding}, and \textit{Focal Convergence}. As depth increases, Phase~2 expands while Phases~1 and~3 remain roughly constant, indicating that additional capacity is primarily used for candidate disambiguation. 

\textbf{Limitations.}
\label{sec:limitations}
Our analysis is restricted to two decoder-only families with dense attention and standard residual connections; extension to MoE, state-space, or non-residual architectures is left to future work. Phase interpretations are based on correlational evidence without causal interventions; activation patching or causal mediation analyses could provide stronger mechanistic grounding. The analysis was conducted using the SlimPajama; validation on additional corpora is left to future work.


\bibliographystyle{plainnat}
\bibliography{references}


\newpage
\appendix

\begin{table}[t]
    \centering
    \caption{Structure of the TunedLens train and validation subsets, extracted from SlimPajama.}
    \resizebox{.4\textwidth}{!}{
    \begin{tabular}{ccc}
        \toprule
        Data Source & Train Tokens & Eval Tokens \\
        \midrule
        CommonCrawl & 130.33M & 10.83M \\
        C4 & 65.72M & 5.20M \\
        ArXiv & 11.52M & 0.99M \\
        GitHub & 9.66M & 0.92M \\
        StackExchange & 7.31M & 0.54M \\
        Wikipedia & 1.95M & 0.18M \\
        \midrule
        Total & 226.49M & 18.66M \\
        \bottomrule
    \end{tabular}
    }
    \label{tab:DatasetStatsTable}
\end{table} 

\begin{figure}[t]
    \centering
    \includegraphics[width=\linewidth]{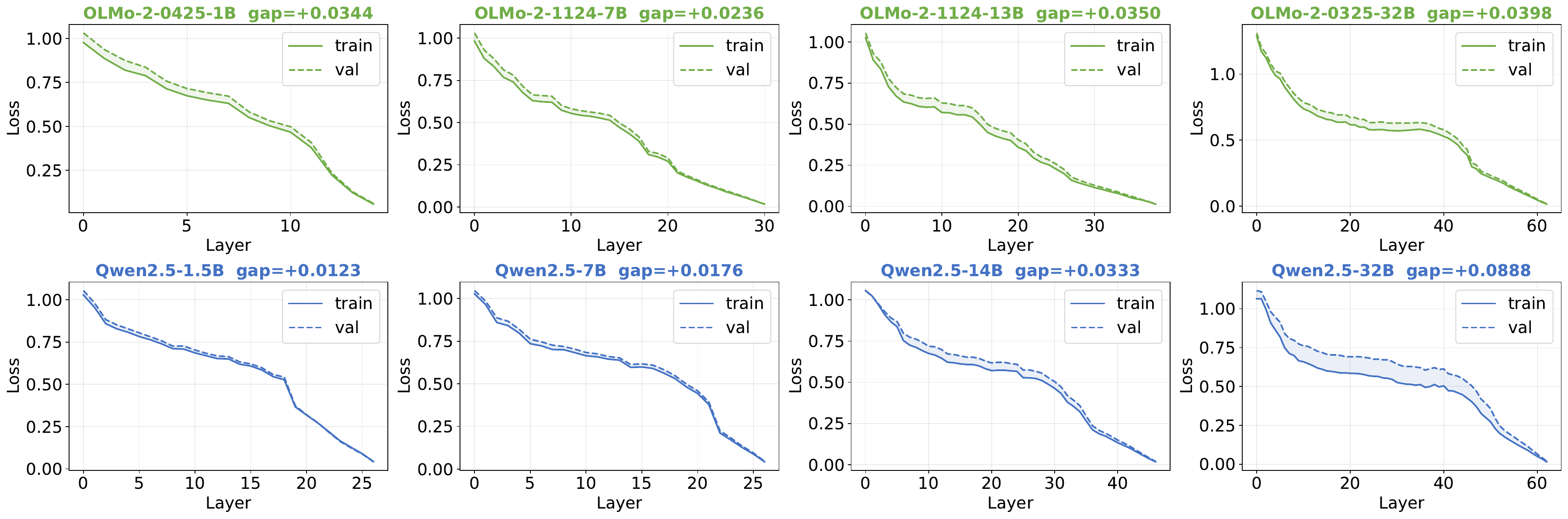}
    \caption{Final per-layer training and validation losses across model scales and families. The low gap indicates no overfitting of trained lenses.}
    \label{fig:lens_train_losses}
\end{figure}

\section{Representation Lens Training and Experiments Setup}
\label{app:lens_training}
 
This appendix documents the training procedure and dataset composition used
to obtain the representation lenses analyzed throughout the paper. All lenses
are trained from scratch on the same corpus and under identical
hyperparameter configurations, ensuring that any observed variation in the
estimated predictive subspaces across model families and scales reflects
genuine model-level differences rather than artifacts of the training
procedure. To train lenses, we used the official TunedLens implementation, which can be found on GitHub\footnote{https://github.com/AlignmentResearch/tuned-lens}. We additionally provide empirical evidence that the trained lenses
do not overfit to the training distribution, supporting the validity of the
geometric analyses conducted on held-out data.

\textbf{Dataset.} All TunedLens are trained on a subset of the SlimPajama training corpus\footnote{https://huggingface.co/datasets/DKYoon/SlimPajama-6B}, comprising a total of 226.49M tokens drawn from six heterogeneous sources: CommonCrawl, C4, ArXiv, GitHub, StackExchange, and Wikipedia. The precise composition of both the training and evaluation subsets is reported in Table~\ref{tab:DatasetStatsTable}. This diversity of source domains, spanning natural language text, scientific writing, source code, and structured web content, encourages the lens to recover predictive structure that is not specific to a narrow distributional regime, and supports the generalization of our geometric findings across input types. For all input-dependent experiments reported in this paper, we employ a held-out subset of the SlimPajama validation set, with per-source composition reported in Table~\ref{tab:DatasetStatsTable}.

\textbf{Optimization.} Training is performed using the AdamW optimizer~\cite{adamw} with a learning rate of $10^{-3}$ and linear decay. Each run consists of 500 optimizer steps, of which the first 50 constitute a linear warmup phase. Each optimizer step processes a batch of $2^{18}$ tokens. All lenses, across all model families and scales, are trained under identical hyperparameter configurations to ensure that differences in the estimated predictive subspaces reflect genuine model-level variation rather than artifacts of the training procedure.

\textbf{Generalization of trained lenses.}
To verify that the trained lenses do not overfit to the training distribution, we track the per-layer distillation loss on both the training and validation subsets throughout training. 
Figure~\ref{fig:lens_train_losses} reports the final per-layer training and validation losses for all models examined. Across all layers and model families, training and validation losses are in close agreement, with no systematic gap indicative of overfitting. 
This consistency confirms that the lens parameters have converged to solutions that generalize beyond the training corpus, and that the predictive subspaces they identify reflect genuine structural properties of the model's residual stream. The geometric analyses reported in the main paper are therefore conducted on the held-out validation subset with full confidence in the reliability of the learned lens parameters.

\textbf{Computational Resources.} All training and evaluation experiments 
were conducted on a Linux server equipped with two 
\textsc{Intel Xeon Gold 6430} CPUs, 1\,TB of RAM, and two 
\textsc{NVIDIA L40S} GPUs with 48\,GB memory each.

\textbf{Execution Time.} Under the specified training settings and hardware 
configuration, lens training required from a few hours for 1B models up to 
approximately four days for 32B models. Evaluation experiments required from 
a few hours for 1B models up to approximately one day and a half for 32B models.

\begin{figure}[t]
    \centering
    \includegraphics[width=\linewidth]{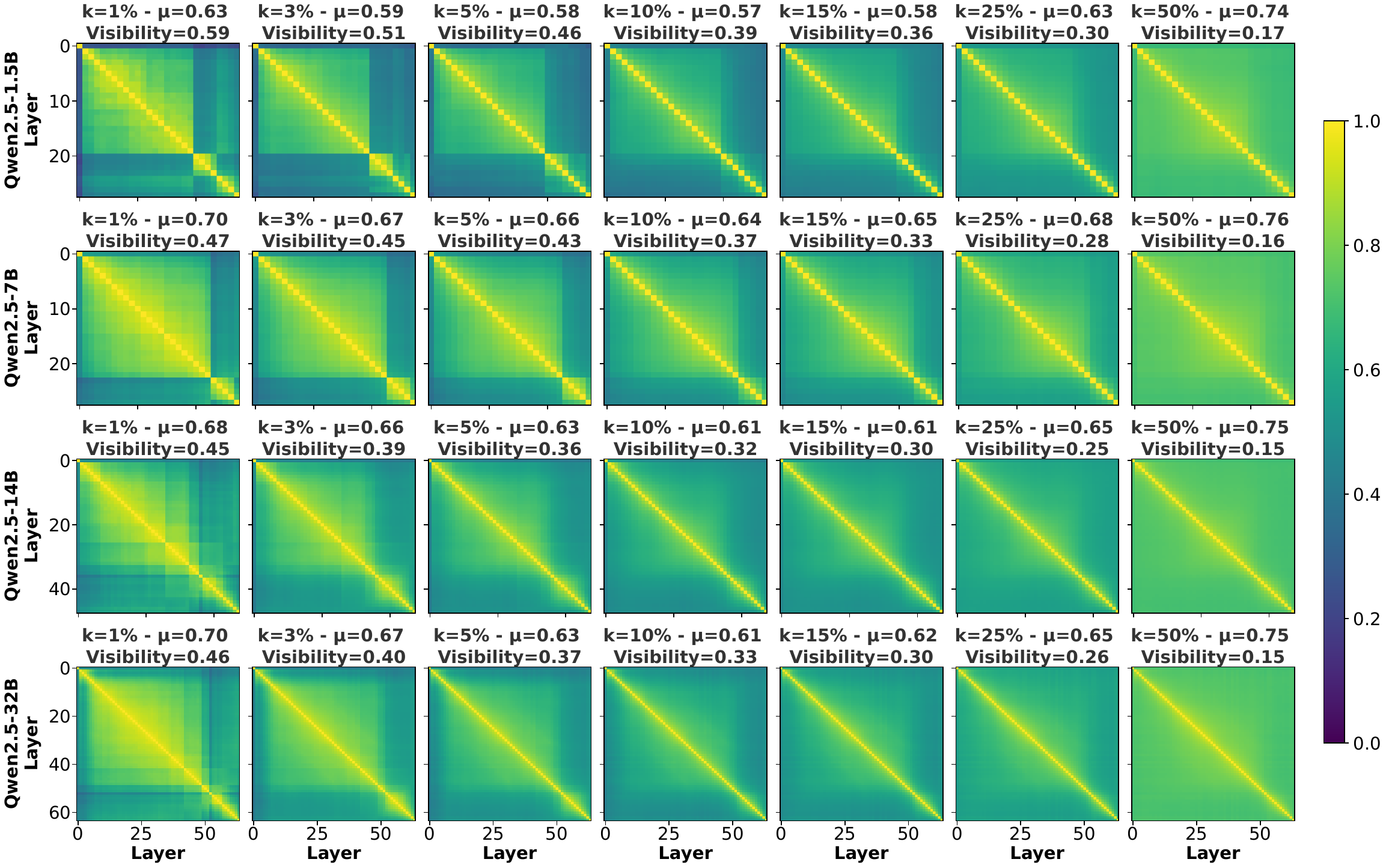}
    \caption{Pairwise RSS similarity matrices $\mathcal{M}^{(k)}$ for all
    models in the \textsc{Qwen2.5} family across truncation levels $k \in
    \{1, 3, 5, 10, 15, 25, 35, 50\}\%$ of $d$. Each row corresponds to a
    model; each column to a truncation level. Geometric contrast decreases
    monotonically with $k$ across all sizes.}
    \label{fig:sim_matrices_qwen}
\end{figure}

\begin{figure}[t]
    \centering
    \includegraphics[width=0.95\linewidth]{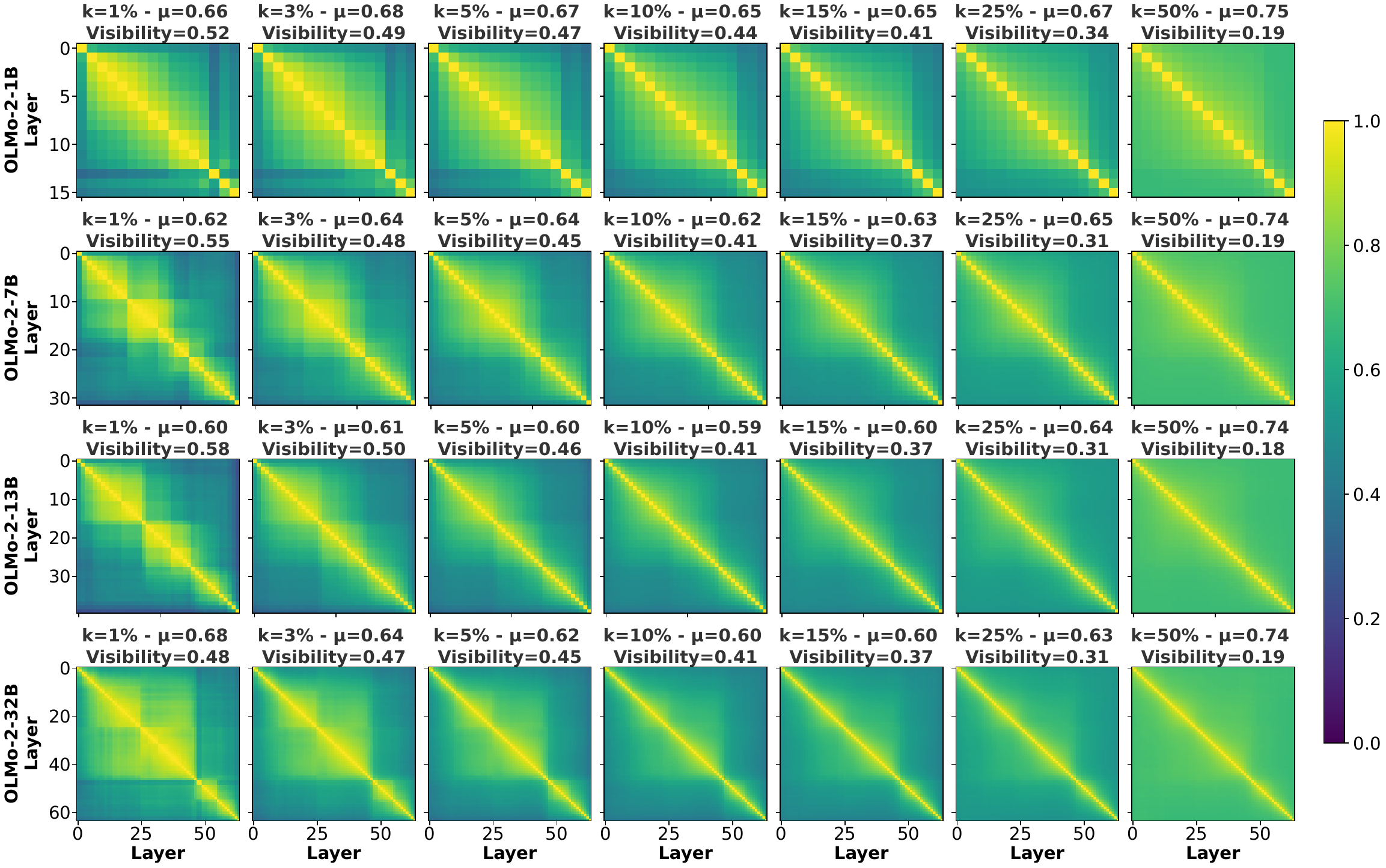}
    \caption{Pairwise RSS similarity matrices $\mathcal{M}^{(k)}$ for all
    models in the \textsc{OLMo2} family across truncation levels $k \in
    \{1, 3, 5, 10, 15, 25, 35, 50\}\%$ of $d$. The same monotonic decay in
    $\mathrm{Visibility@K}$ is observed consistently across all model sizes,
    in agreement with the \textsc{Qwen2.5} family.}
    \label{fig:sim_matrices_olmo}
\end{figure}

\section{Pairwise RSS Similarity Matrices Across Truncation Levels}
\label{app:sim_matrices}

Section~\ref{sec:resolution} introduces the pairwise RSS similarity matrix
$\mathcal{M}^{(k)}$, whose off-diagonal spread, quantified by
$\mathrm{Visibility@K}$, decreases monotonically as the truncation level
$k$ increases. The main text illustrates this behavior for
\textsc{OLMo2-32B} (Figure~\ref{fig:sim_matrices}). Here we provide the
complete set of pairwise similarity matrices across all truncation levels $k
\in \{1, 3, 5, 10, 15, 25, 35, 50\}\%$ of $d$ for all models in both families examined.

Figures~\ref{fig:sim_matrices_qwen} and~\ref{fig:sim_matrices_olmo} report
these matrices for the \textsc{Qwen2.5} and \textsc{OLMo2} families,
respectively. Each row corresponds to a model and each column to a truncation
level. Several observations are consistent across both families. At low
values of $k$, the matrices exhibit high geometric contrast: off-diagonal
entries span a wide range, reflecting well-separated subspaces and a rich
layer-wise structure that is clearly visible in the similarity patterns.
As $k$ increases, the matrices become progressively more uniform, with
off-diagonal entries converging toward higher values and the overall contrast
declining. In the limit $k \to d$, all subspaces approach $\mathbb{R}^d$
and pairwise similarity collapses toward $1$ uniformly, causing
$\mathrm{Visibility@K}$ to approach zero. This behavior is consistent across
model sizes within each family, confirming that the trade-off between
geometric discriminability and spectral energy retention is a structural
property of the representation lens geometry rather than a scale-dependent
artifact.

\begin{figure}[t]
    \centering
    \includegraphics[width=0.95\linewidth]{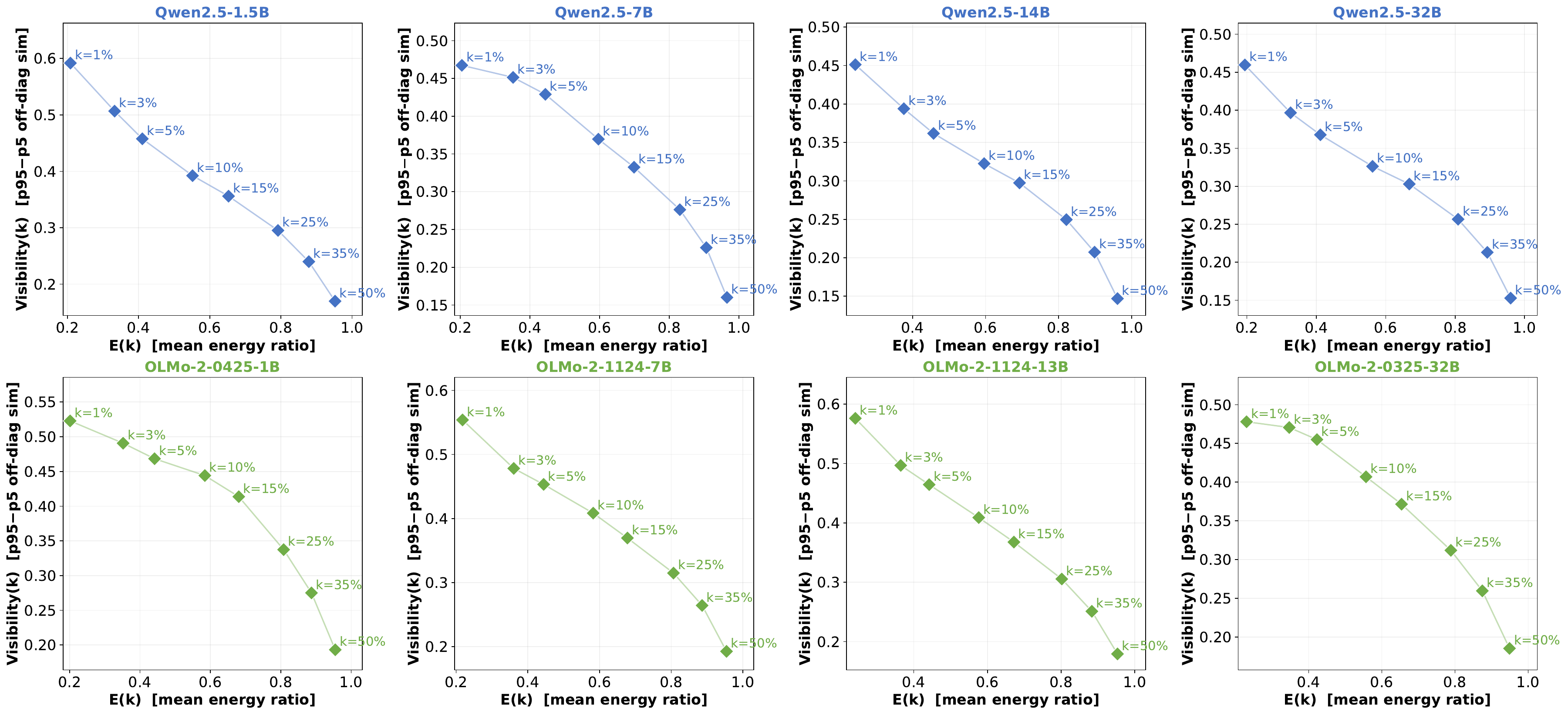}
    \caption{Pareto frontier between $\mathrm{Visibility@K}$ and $E^{(k)}$
    for all models in the \textsc{Qwen2.5} and \textsc{OLMo2} families.
    Each point corresponds to a truncation level $k \in \{1, 3, 5, 10, 15,
    25, 35, 50\}\%$ of $d$. The intermediate regime $k \in \{5\%, 10\%,
    15\%\}$ consistently lies at the region of highest curvature across all
    model sizes and both families, offering the most favorable trade-off
    between spectral fidelity and geometric discriminability.}
    \label{fig:pareto_full}
\end{figure}

\section{Pareto Frontier Between Visibility@K and Spectral Energy}
\label{app:pareto}

Section~\ref{sec:resolution} establishes that varying the truncation level
$k$ traces a Pareto frontier between geometric discriminability, as measured
by $\mathrm{Visibility@K}$, and spectral energy retention, as measured by
$E^{(k)}$. The main text illustrates this frontier for a representative
subset of models (Figure~\ref{fig:pareto_frontier}). Here we provide the
complete Pareto curves for all models in both the \textsc{Qwen2.5} and
\textsc{OLMo2} families.

Figure~\ref{fig:pareto_full} reports $\mathrm{Visibility@K}$ against
$E^{(k)}$ for each model and truncation level $k \in \{1, 3, 5, 10, 15,
25, 35, 50\}\%$ of $d$. Across all models and both families, the frontier
exhibits the same qualitative structure: no single value of $k$ dominates on
both objectives simultaneously, and the curve is strictly monotone in the
sense that increasing $k$ always improves $E^{(k)}$ at the cost of reducing
$\mathrm{Visibility@K}$. The intermediate regime $k \in \{5\%, 10\%, 15\%\}$
consistently corresponds to the region of highest curvature of the frontier,
providing a favorable balance between fidelity and geometric discriminability
across all model sizes and both architectural families. This structural
regularity supports the use of the intermediate regime as the primary
resolution scale for phase detection.

\begin{figure}[t]
    \centering
    \includegraphics[width=\linewidth]{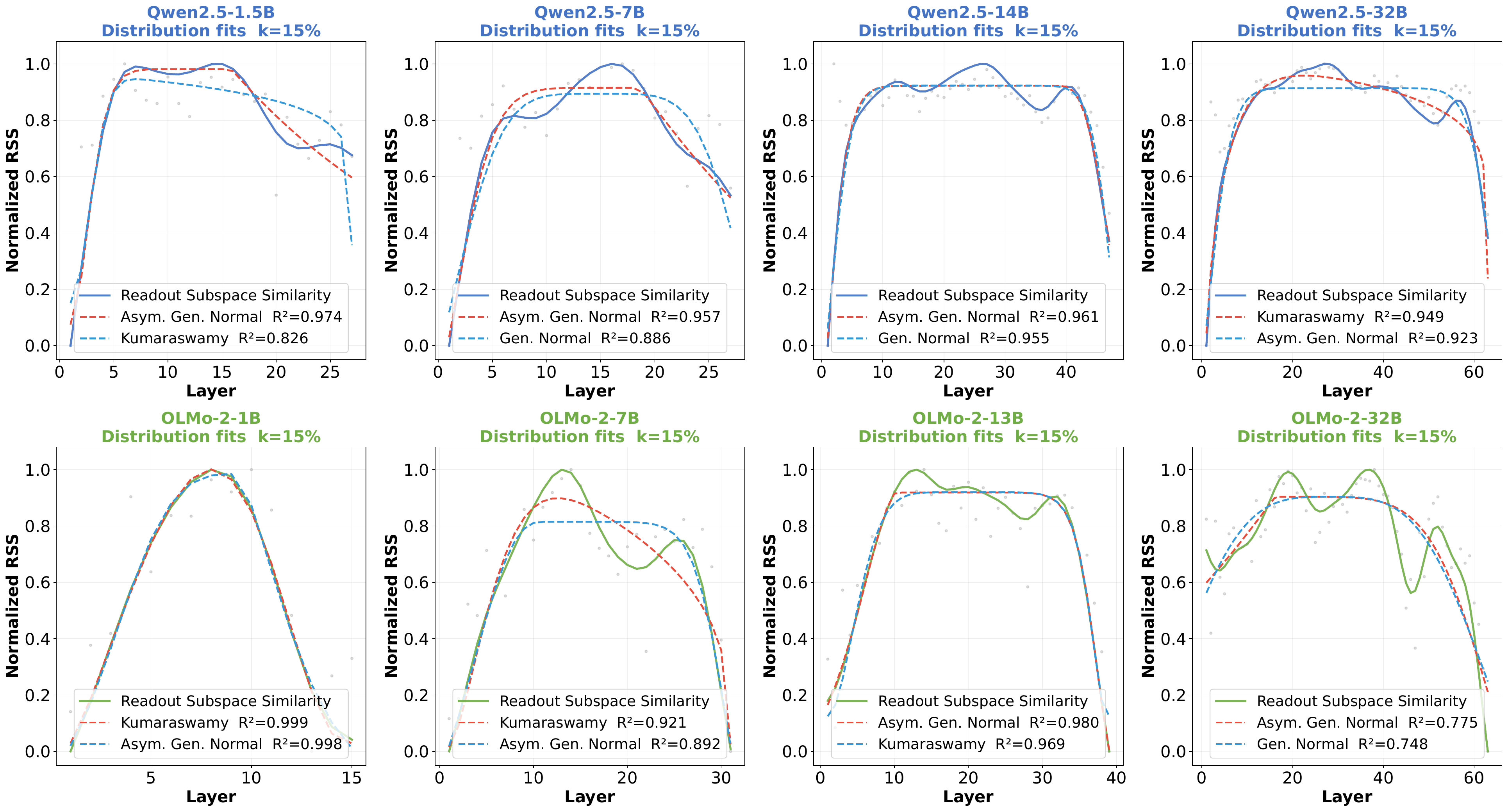}
    \caption{Distribution fits (left panels) and double-chord breakpoint
    detection (right panels) on the RSS profile $\{s_\ell^{(k=15\%)}\}$ for
    all models in the \textsc{Qwen2.5} and \textsc{OLMo2} families. Left panels show the
    normalized similarity profile overlaid with fits from the four parametric
    families (Beta, GNorm, AGNorm, Kumaraswamy). Right panels show the
    smoothed profile with identified breakpoints $\hat{b}_1$ and $\hat{b}_2$
    and the corresponding chord constructions. Unimodal families achieve
    consistently high $R^2$ across all model sizes; the more flexible
    families do not systematically improve the fit.}
    \label{fig:fits_bp_all}
\end{figure}

\section{Unimodality Validation and Phase Detection Across All Models}
\label{app:unimodality_phase_detection}

Section~\ref{sec:resolution} introduces the unimodality assumption for the RSS
similarity profiles $\{s_\ell^{(k)}\}$ and validates it by fitting four
parametric families (Beta, GNorm, AGNorm, and Kumaraswamy) to the
normalized profiles.
The main text illustrates both analyses for \textsc{Qwen2.5-14B} and
\textsc{OLMo2-13B} at $k = 15\%$ (Figure~\ref{fig:example_breakpoints}).
Here we provide the complete set of distribution fits and breakpoint
detections for all models in both families at the same truncation level.

Figure~\ref{fig:fits_bp_all} report, for each
model in the \textsc{Qwen2.5} and \textsc{OLMo2} families, the
distribution fits (left panels) and the double-chord breakpoint detection
(right panels) on the RSS profile $\{s_\ell^{(k=15\%)}\}$. Each left panel
displays the normalized similarity profile overlaid with the best-fitting
curve from each of the four parametric families; each right panel shows the
smoothed profile with the identified breakpoints $\hat{b}_1$ and $\hat{b}_2$
and the corresponding chord constructions used for their detection.

Across all models and both families, the unimodal families achieve high
goodness of fit, capturing the global rise--plateau--descent structure of the
profiles. The more flexible families (Beta and Kumaraswamy) do not
systematically outperform the unimodal ones (GNorm and AGNorm), confirming
that the additional expressivity is not required to account for the observed
profile structure and that the unimodal assumption does not discard relevant
information. Local deviations from a strictly unimodal shape are present in
some models, particularly in the transition regions; the chord-based detection
procedure is robust to these deviations because of operating on the global
geometry of each half-profile rather than on local derivatives. The identified breakpoints
$\hat{b}_1$ and $\hat{b}_2$ are consistent in their structural placement
across all models: $\hat{b}_1$ falls reliably in the first half of the
network and $\hat{b}_2$ in the second half, confirming that the tripartite
decomposition is a stable property of the forward pass geometry across
architectures and scales.

\subsection{Breakpoints detection.}
Given the unimodal similarity profile $\{s_\ell^{(k)}\}_{\ell=1}^{L-1}$, we identify two breakpoints $b_1 < b_2$ that partition depth into three phases via a two-stage Kneedle criterion~\citep{satopaa2011finding}, applied separately to each half of the profile. This chord-based method is assumption-light, avoids derivatives, and is robust to local deviations from unimodality (Figure~\ref{fig:example_breakpoints}), as it relies on global geometry. All computations use a Gaussian-smoothed profile with $\sigma = 2$. Let $\tilde{s}_\ell$ be the smoothed profile and $D(\ell, \overline{AB})$ the distance from $(\ell, 
\tilde{s}_\ell)$ to segment $\overline{AB}$. The breakpoints are:
\[
\begin{aligned}
b_1 &= \underset{\ell \in [1,m]}{\arg\max}\;
D\!\left(\ell,\,\overline{(1,\tilde{s}_1)(m,\tilde{s}_m)}\right), \\
b_2 &= \underset{\ell \in [m,L-1]}{\arg\max}\;
D\!\left(\ell,\,\overline{(m,\tilde{s}_m)(L{-}1,\tilde{s}_{L-1})}\right).
\end{aligned}
\]

where $L$ is the number of layers and $m = \lfloor L/2 \rfloor$. The midpoint split follows from unimodality, placing one transition in each half. Breakpoint estimates may vary by $1$--$3$ layers across values of $k$ due 
to residual noise. We aggregate over the intermediate regime 
$k \in \{5\%, 10\%, 15\%\}$ via:
\[
\hat{b}_1 = \mathrm{mode}\bigl\{b_1^{(k)} : k \in \{5, 10, 15\}\bigr\}, 
\qquad
\hat{b}_2 = \mathrm{mode}\bigl\{b_2^{(k)} : k \in \{5, 10, 15\}\bigr\},
\]
using the median when all three estimates are distinct. Robustness to other $k$ values is discussed in Section~\ref{sec:results}.

\begin{figure}[t]
    \centering
    \includegraphics[width=\linewidth]{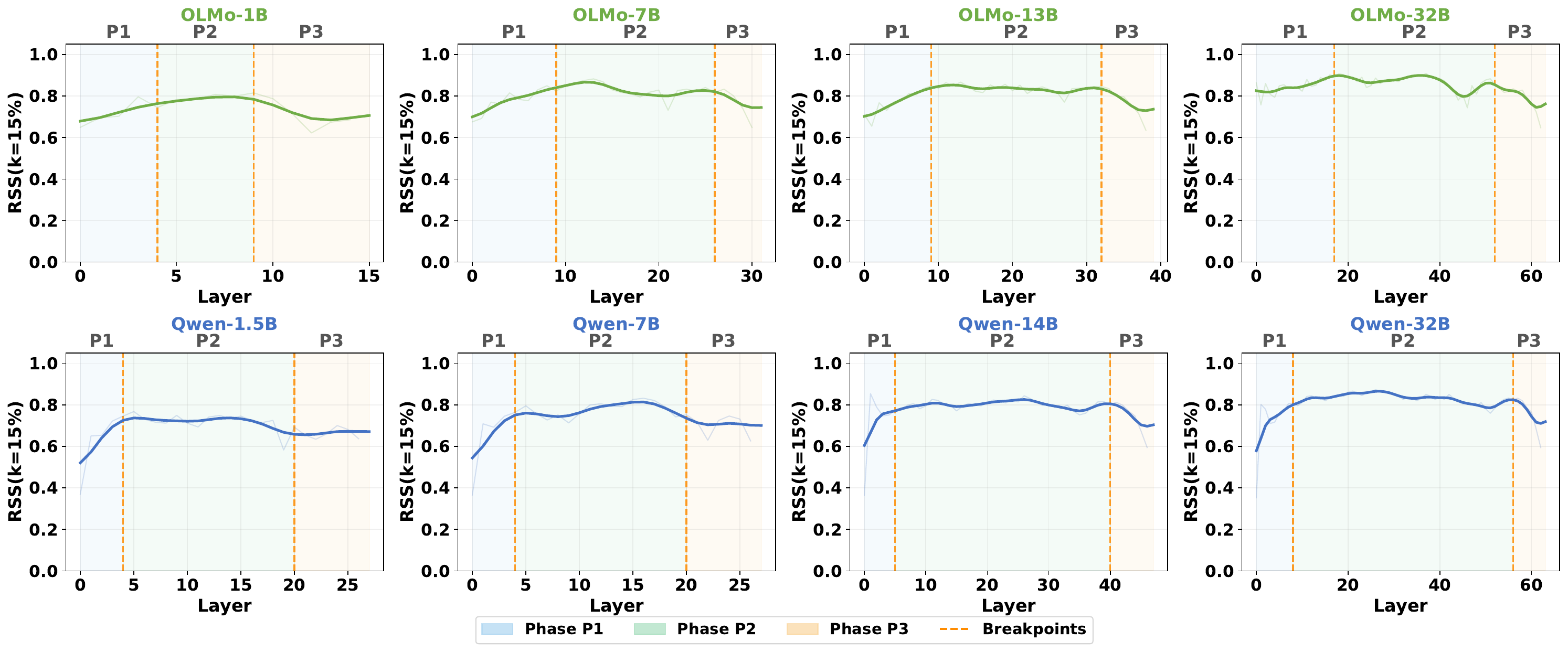}
    \caption{RSS profiles $s_\ell^{(k=15\%)}$ for all eight models across
    both families, with consensus breakpoints $\hat{b}_1$ and $\hat{b}_2$
    indicated. The rise--plateau--descent structure and the placement of
    phase boundaries are consistent across all model sizes and both
    architectural families.}
    \label{fig:phase_overview_full}
\end{figure}

\section{Geometric Characterization of the Three Phases: Full Results}
\label{app:geometric_characterization}

\begin{figure}[t]
    \centering
    \includegraphics[width=\linewidth]{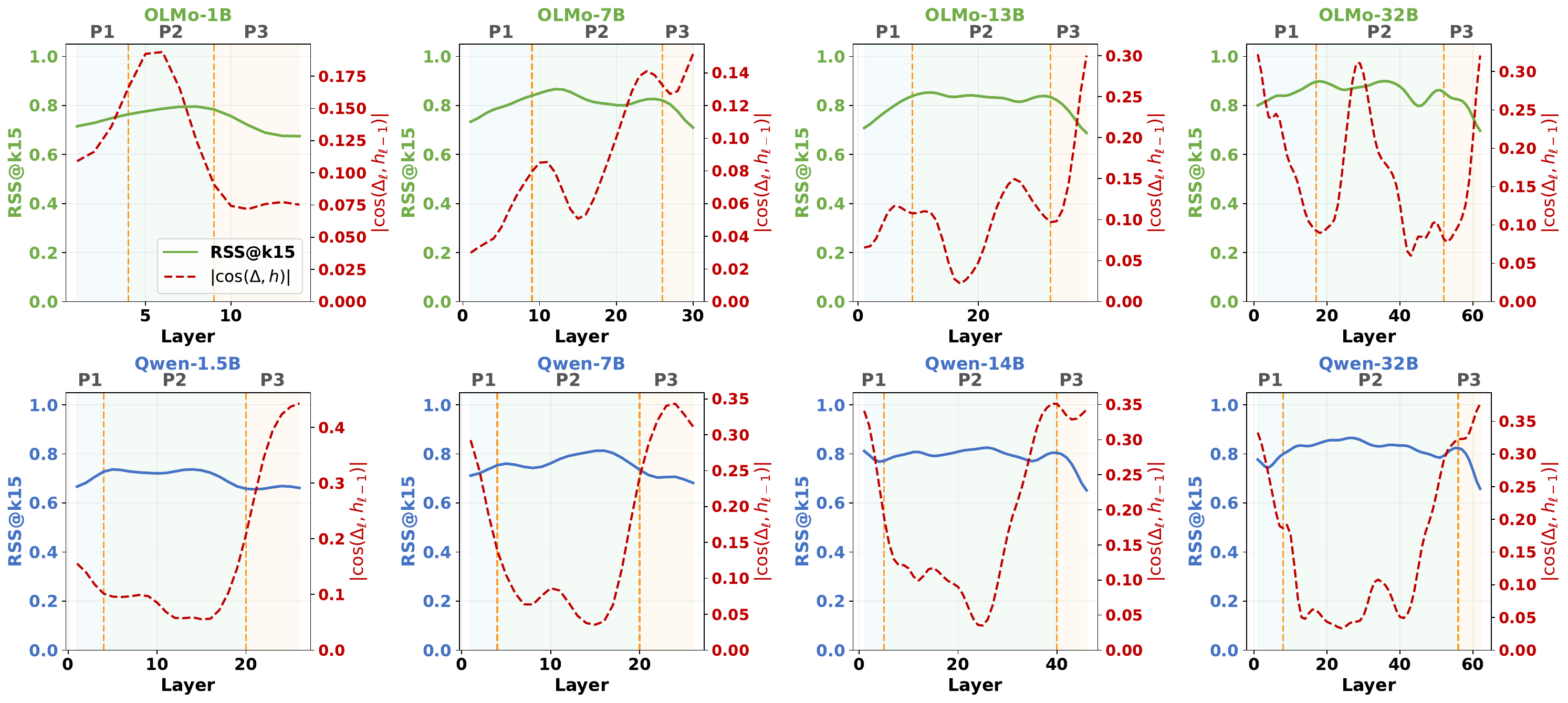}
    \caption{RSS profile $s_\ell^{(k=15\%)}$ (solid, left axis) and
    $|\cos(\Delta_\ell, h_{\ell-1})|$ (dashed red, right axis) overlaid
    for all eight models across both families. High RSS $\leftrightarrow$
    low cosine (Phase~2); low RSS $\leftrightarrow$ high cosine (Phase~3).
    The anti-correlation is consistent across all models and scales.}
    \label{fig:bridge_full}
\end{figure}

\begin{figure}[t]
    \centering
    \includegraphics[width=\linewidth]{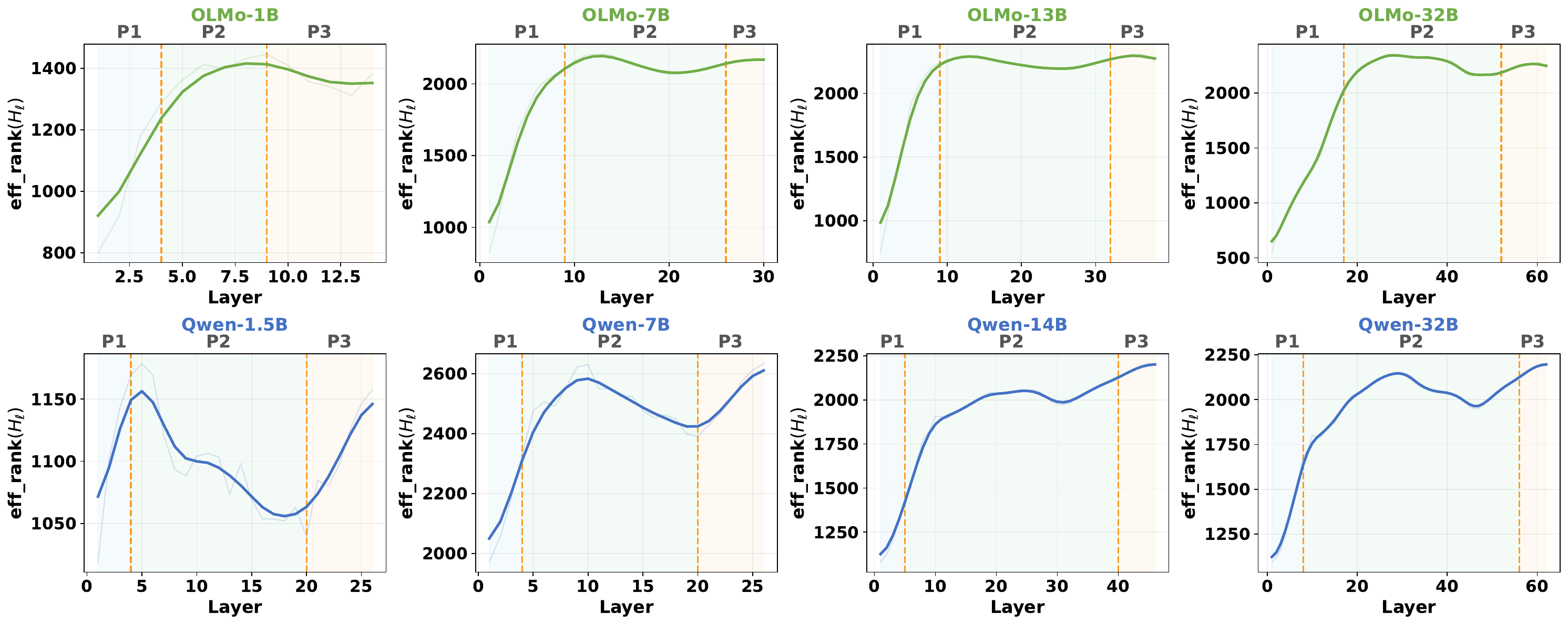}
    \caption{Effective rank of hidden states $H_\ell$ averaged across 16M
    tokens for all eight models. Rank increases through Phase~1, peaks near
    $\hat{b}_1$, stabilises in Phase~2, and remains approximately constant
    through Phase~3. \textsc{Qwen2.5-1.5B} exhibits reduced layer-wise
    variance consistent with its constrained capacity.}
    \label{fig:rank_saturation_full}
\end{figure}

\begin{figure}[t]
    \centering
    \includegraphics[width=\linewidth]{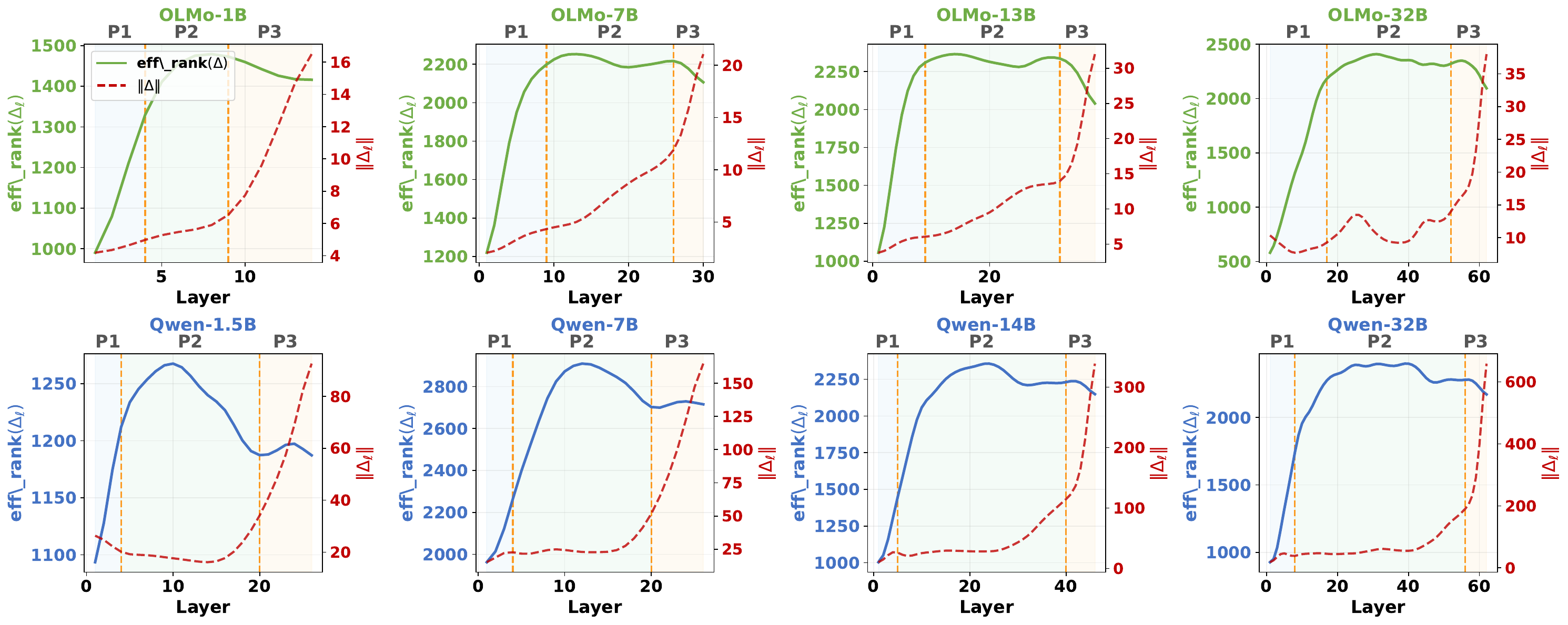}
    \caption{Effective rank of the total update $\Delta_\ell$ (solid) and
    its norm $\|\Delta_\ell\|_2$ (dashed red) for all eight models. Update
    rank peaks in Phase~2 and declines in Phase~3 (7/8 models); update norm
    increases monotonically with sharp acceleration in Phase~3 (up to
    $10\times$), indicating concentration on few high-energy directions.}
    \label{fig:delta_collapse_full}
\end{figure}

Section~\ref{sec:results_characterization} reports the geometric
characterization of the three phases for a representative subset of four
models. Here we provide the complete results for all eight models across
both families (\textsc{Qwen2.5} and \textsc{OLMo2}, 1B--32B parameters),
covering the near-orthogonality of updates and the effective rank analysis
of hidden states and layer updates.

\paragraph{RSS profiles and consensus breakpoints.}
Figure~\ref{fig:phase_overview_full} extends Figure~\ref{fig:phase_overview}
to all eight models, showing the RSS profile $s_\ell^{(k=15\%)}$ with
consensus breakpoints $\hat{b}_1$ and $\hat{b}_2$ across normalised depth.
The tripartite structure is consistent across all models and both families:
Phase~1 exhibits a rising profile as the predictive geometry stabilises,
Phase~2 maintains a near-plateau of high subspace similarity, and Phase~3
shows a decline as the predictive subspace reorients toward the unembedding
direction for decoding. The breakpoints are reliably placed in the first and
second halves of the network respectively, confirming the structural
regularity reported in Section~\ref{sec:results_unimodal}.

\paragraph{Near-orthogonality of updates.}
Figure~\ref{fig:bridge_full} extends Figure~\ref{fig:bridge} to all eight
models, overlaying the RSS profile $s_\ell^{(k=15\%)}$ and the cosine
similarity $|\cos(\Delta_\ell, h_{\ell-1})|$ across normalised depth. The
anti-correlation between the two quantities is consistent across all models
and both families: high RSS aligns with low cosine in Phase~2, and low RSS
with high cosine in Phase~3. In all models, $|\cos(\Delta_\ell,
h_{\ell-1})|$ is lowest in Phase~2 (7/8 models) and highest in Phase~3
(8/8 models), confirming that updates remain nearly orthogonal to the
residual stream throughout the forward pass and that phases differ in the
geometric effect of these updates on the predictive subspace rather than in
their alignment with the existing residual direction.

\paragraph{Effective rank of hidden states and layer updates.}
Figures~\ref{fig:rank_saturation_full} and~\ref{fig:delta_collapse_full}
extend Figures~\ref{fig:rank_saturation} and~\ref{fig:delta_collapse} to
all eight models. The effective rank of hidden states
(Figure~\ref{fig:rank_saturation_full}) increases throughout Phase~1 and
peaks near or shortly after $\hat{b}_1$ across all models, then enters a
low-variance regime in Phase~2 consistent with a stabilised predictive
geometry. The exception is \textsc{Qwen2.5-1.5B}, which shows a modest
dip of $\approx 80$ dimensions in Phase~2 and generally lower layer-wise
rank variance, which we attribute to its constrained capacity ($d = 1536$,
27 layers). The effective rank and norm of layer updates
(Figure~\ref{fig:delta_collapse_full}) follow a consistent pattern: update
rank grows through Phases~1--2, peaks in Phase~2 (8/8 models), then
declines in Phase~3 (7/8 models), while $\|\Delta_\ell\|$ increases
monotonically with sharp acceleration in Phase~3 (up to $10\times$),
indicating concentration of update energy onto a shrinking set of
high-energy directions. These patterns are qualitatively uniform across
model sizes and both architectural families, confirming that the geometric
signatures of the three phases identified in the main text are not
artifacts of the selected subset of models.


\newpage

\end{document}